\documentclass[10pt,twocolumn,letterpaper]{article}

\usepackage[pagenumbers]{cvpr} 

\usepackage{graphicx}
\usepackage{amsmath}
\usepackage{amssymb}
\usepackage{booktabs}
\usepackage{epsfig}
\usepackage{amsmath}
\usepackage{amssymb}
\usepackage{booktabs}
\usepackage{tabulary}
\usepackage{multirow}
\usepackage{overpic}
\usepackage{xspace}
\usepackage{bbm}
\usepackage{xcolor}         
\usepackage{microtype}
\usepackage{amsmath,amsfonts,bbm}
\usepackage{multirow}
\usepackage{epsfig}
\usepackage{times,amsmath,amssymb,booktabs,tabulary,multirow,overpic,xcolor}
\usepackage{mathtools}
\usepackage{wrapfig}
\usepackage{nicefrac}       
\usepackage{microtype}      
\usepackage{enumitem}
\usepackage{bbm}

\usepackage[pagebackref=false, breaklinks=true, letterpaper=true, colorlinks,
            citecolor=citecolor, linkcolor=linkcolor, bookmarks=false]{hyperref}
\definecolor{citecolor}{HTML}{0071BC}
\definecolor{linkcolor}{HTML}{ED1C24}

\renewcommand{\paragraph}[1]{\vspace{1.25mm}\noindent\textbf{#1}}

\usepackage[ruled]{algorithm2e}
\usepackage{algorithmic}

\graphicspath{ {./figs/} }
\usepackage{floatrow}
\usepackage{adjustbox}
\newfloatcommand{capbtabbox}{table}[][\FBwidth]

\usepackage[capitalize]{cleveref}
\crefname{section}{Sec.}{Secs.}
\Crefname{section}{Section}{Sections}
\Crefname{table}{Table}{Tables}
\crefname{table}{Tab.}{Tabs.}

\def\cvprPaperID{000} 
\def\confName{ICCV}
\def\confYear{2023}

\usepackage{multirow}
\usepackage{cuted}

\newcommand{\vct}[1]{\boldsymbol{#1}} 
\newcommand{\mat}[1]{\boldsymbol{#1}} 

\newcommand{\methodname}{{AdaTrans}\xspace}

\newcommand{\app}{\raise.17ex\hbox{$\scriptstyle\sim$}}

\newlength\savewidth

\usepackage{subfig}
\usepackage{overpic}

\begin{document}

\title{
Adaptive Nonlinear Latent Transformation for Conditional Face Editing
}

\author{Zhizhong Huang$^{1}$\qquad Siteng Ma$^{1}$\qquad Junping Zhang$^{1}$\qquad Hongming Shan$^{2,3}$\thanks{Corresponding author}
\\
$^{1}$ Shanghai Key Lab of Intelligent Information Processing, School of Computer Science,\\
Fudan University, Shanghai 200433, China\\
$^{2}$ Institute of Science and Technology for Brain-inspired Intelligence and MOE Frontiers Center \\for Brain Science,  Fudan University, Shanghai 200433, China\\
$^{3}$ Shanghai Center for Brain Science and Brain-inspired Technology, Shanghai 200031, China\\
{\tt\small \{zzhuang19, stma21, jpzhang, hmshan\}@fudan.edu.cn}
}

\maketitle
\thispagestyle{empty}
\begin{strip}
    \vspace{-0.6in}
    \centering
    \begin{minipage}{\textwidth}
        \centering
    \includegraphics[width=0.94\textwidth]{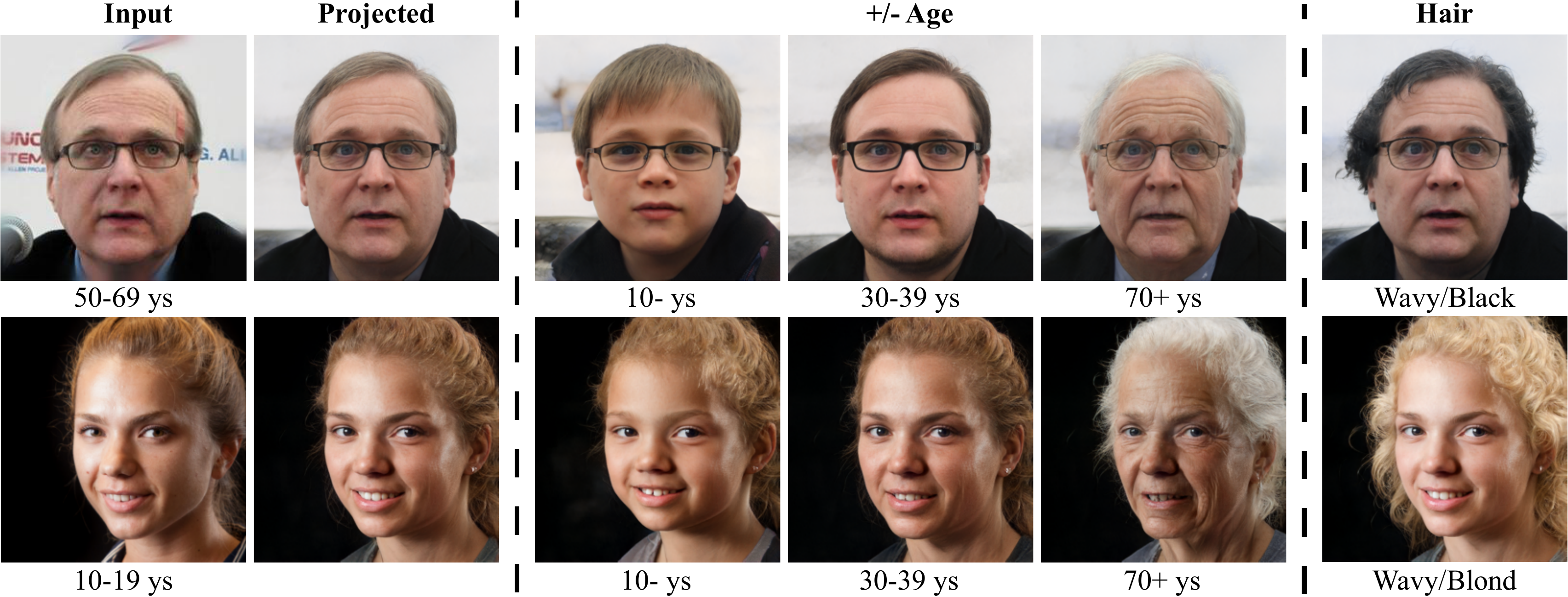}
    \vspace{-0.2cm}
    \captionof{figure}
    {
        Sample results produced by our \methodname. By projecting the images into the latent space of StyleGAN, \methodname can achieve disentangled face editing even when the age gap is extreme large, and manipulate multiple attributes at the same time.
    }
    \label{fig:teaser}
    \end{minipage}
\end{strip} 

\begin{abstract}

Recent works for face editing usually manipulate the latent space of StyleGAN via the linear semantic directions.
However, they usually suffer from the entanglement of facial attributes, need to tune the optimal editing strength, and are limited to binary attributes with strong supervision signals.
This paper proposes a novel adaptive nonlinear latent transformation for disentangled and conditional face editing, termed \methodname.
Specifically, our \methodname divides the manipulation process into several finer steps; \ie, the direction and size at each step are conditioned on both the facial attributes and the latent codes. In this way, \methodname describes an adaptive nonlinear transformation trajectory to manipulate the faces into target attributes while keeping other attributes unchanged.
Then, \methodname leverages a predefined density model to constrain the learned trajectory in the distribution of latent codes by maximizing the likelihood of transformed latent code.
Moreover, we also propose a disentangled learning strategy under a mutual information framework to eliminate the entanglement among attributes, which can further relax the need for labeled data.
Consequently, \methodname enables a controllable face editing with the advantages of disentanglement, flexibility with non-binary attributes, and high fidelity.
Extensive experimental results on various facial attributes demonstrate the qualitative and quantitative effectiveness of the proposed \methodname over existing state-of-the-art methods, especially in the most challenging scenarios with a large age gap and few labeled examples.
The source code is available at \url{https://github.com/Hzzone/AdaTrans}.
\end{abstract}
\section{Introduction}

Face editing aims to render the faces to the target facial attributes such as aging or smiling with high fidelity while keeping other facial attributes unchanged, which has wide applications in entertainment and forensics.
Due to the intrinsic complexity of facial attributes, face editing has attracted growing research interest in recent years. Generative Adversarial Networks (GANs)~\cite{goodfellow2014generative} have shown promising results for face editing in terms of image quality. Earlier works mainly focus on network architectures~\cite{zhu2017unpaired,isola2017image,xu2022transeditor} and loss functions~\cite{park2020contrastive}. With significant improvements in image quality, these methods usually re-train a GAN model for a specific facial attribute~\cite{huang2020pfa} or practical applications~\cite{liu2019stgan}. Unfortunately, due to the difficulties in training a good GAN, they are limited to specific tasks and fail to generalize to high-resolution images.

In recent years, StyleGAN~\cite{karras2019style,karras2020analyzing,karras2021alias} has achieved significant progress in synthesizing photorealistic faces. In particular, the pre-trained StyleGAN generator presents a meaningful intermediate latent space, traversing on which the faces can be semantically manipulated~\cite{shen2020interpreting,yao2021latent,harkonen2020ganspace,shen2021closed,yang2021discovering,abdal2021styleflow,wu2021stylespace,abdal2022clip2stylegan}.
Typically, the faces before editing need to be inverted into the latent space of StyleGAN to obtain the latent codes that can be used to faithfully reconstruct the inputs~\cite{tov2021designing,alaluf2021restyle,richardson2021encoding}. As a result, the latent code is manipulated along certain directions, giving rise to the changes in the corresponding attribute in generated faces. 
The methods to obtain those directions can be roughly categorized as supervised ones and unsupervised ones.
The supervised methods~\cite{shen2020interpreting,yao2021latent,abdal2021styleflow,wu2021stylespace} leverage the labeled data to compute the semantic directions, leading to better controllability in the editing process. For example, InterFaceGAN~\cite{shen2020interpreting} trains a hyperplane in the latent space to separate the examples with binary attributes. Unsupervised methods~\cite{harkonen2020ganspace,shen2021closed,yang2021discovering,abdal2022clip2stylegan} are to discover the interpretable directions using PCA~\cite{harkonen2020ganspace,shen2021closed} or texts~\cite{abdal2022clip2stylegan}. Despite the meaningful transformations, they cannot produce precise user-desired editing without any human annotations.

In summary, most of these methods assume that the binary attributes can be well separated, so they edit the faces by linear interpolation in the latent space.
Although it is sufficient to some degree, for those more complicated scenarios, \eg, with large age gaps, they cannot perform disentangled editing to preserve the unrelated attributes when linear assumption does not hold.
Meanwhile, the users are usually required to manually tune the editing strength for accurate manipulation. Though flexible, the optimal strength varies among different examples. Furthermore, there is a critical yet ignored problem in current literature that the latent codes could be over-manipulated, \ie falling out of the latent space, which inevitably harms the quality of the edited face.

In this paper, we propose a novel framework for conditional face editing, termed \methodname, to address these issues in the following aspects.
\emph{First}, instead of manually manipulating the latents with fixed directions, we propose an adaptive nonlinear transformation strategy that dynamically estimates the editing direction and step size, conditioned by the target attributes and transformation trajectory. Such a strategy can handle various attributes at the same time for conditional multi-attribute editing, by only changing the target attributes while keeping others unchanged.
\emph{Second}, we propose to maximize the likelihood of edited latent codes,  regularize the transformed trajectory in the distribution of latent space, and hence improve the fidelity of edited faces predicted by a pretrained generator.
\emph{Last}, we propose a disentangled learning strategy under a mutual information framework, attenuating the entanglement between attributes and relaxing the need for labeled data in supervised face editing methods.
The merits of \methodname are disentanglement, high fidelity, controllability, and flexibility. The sample results are shown in Fig.~\ref{fig:teaser}.

The contributions are summarized as follows:
\begin{itemize}
[leftmargin=*]
\setlength{\itemsep}{0pt} 
\setlength{\parskip}{0pt}%
\setlength{\topsep}{0pt}%
\item We present \methodname, a novel face editing method that explores an adaptive nonlinear transformation for disentangled and multi-attribute face editing. 
\item We propose a novel density regularization term, which can encourage an in-distribution transformation in the latent space, without harming fidelity.
\item We further show a disentangled learning strategy, which can eliminate the entanglement between attributes and relax the need for labeled data.
\item Experimental results on various facial attributes demonstrate the effectiveness of \methodname both quantitatively and qualitatively. In particular, \methodname can produce disentangled editing, even with extremely large age gaps or few labeled data.
\end{itemize}
\section{Related Work}
\paragraph{Generative adversarial networks.}
Generative Adversarial Networks~(GANs) describe a competition between the generator and discriminator, where the generator maps the random noise~(\eg Gaussian) to the complicated data distribution~(\eg image), and the discriminator tries to distinguish the true/generated data. Various works have made significant progress in synthesizing photorealistic faces from different aspects such as loss functions~\cite{mao2017least,arjovsky2017wasserstein} and architectures~\cite{karras2017progressive,karras2019style,brock2018large}. In particular, StyleGAN~\cite{karras2019style,karras2020analyzing,karras2021alias} presents a meaningful intermediate latent space $\mathcal{W}$ which is better disentangled than a standard Gaussian latent space $\mathcal{Z}$. To utilize a well pre-trained StyleGAN generator, the faces are first inverted to $\mathcal{W}$ to obtain the latent codes that can be used to faithfully reconstruct the input images~\cite{zhu2020domain,tov2021designing,alaluf2021restyle,richardson2021encoding,hu2022style}. Consequently, semantical face editing can be achieved by manipulating the latent space of GANs, which is fed into the generator to obtain the manipulated faces.

\begin{figure*}[t]
    \centering
    \includegraphics[width=1.00\linewidth]{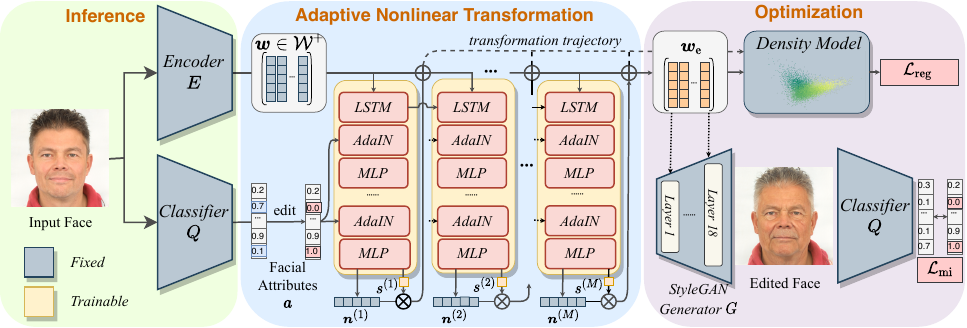}
    \caption{
    Illustration of the proposed \methodname.
    Given an input face, the pre-trained classifier predicts its attributes and the encoder inverts it to the latent space of StyleGAN generator. By changing the facial attributes, our proposed \methodname can manipulate the faces in the latent space for adaptive nonlinear transformation without tuning the strength. The proposed density regularization term can encourage the in-distribution edited latent codes and the disentangled learning strategy can attenuate the entanglement between attributes. The merits of \methodname are flexible, disentangled, controllable, and nonlinear.    \label{fig:framework}
    }
\end{figure*}

\paragraph{Face editing in GANs.}
The prior literature on face editing can be roughly split into supervised and unsupervised methods.
The supervised methods~\cite{shen2020interpreting,abdal2021styleflow,yao2021latent,khrulkov2021latent,yang2021discovering,zhuang2021enjoy,wu2021stylespace} typically employ the human annotations with particular facial attributes or a pre-trained classifier to identify how to manipulate the faces in the latent space.
InterFaceGAN~\cite{shen2020interpreting} uses the hyperplane separating latent codes with binary attributes to carry out the linear editing.
StyleFlow~\cite{abdal2021styleflow} remaps the latent codes to Gaussian noise with normalizing flows and then samples the new latent codes conditioned by the target attributes and original noise.
Latent Transformer~\cite{yao2021latent} trains a transformation network for specified attributes under the supervision of other attributes.
Differently, the unsupervised methods~\cite{jahanian2019steerability,voynov2020unsupervised,harkonen2020ganspace,shen2021closed,Image2StyleGANpp,abdal2022clip2stylegan,collins2020editing} do not need the valuable annotated data.
GANSpace~\cite{harkonen2020ganspace} and SeFa~\cite{shen2021closed} find editing directions from  the principal components of the latent space.
Clip2Stylegan~\cite{abdal2022clip2stylegan} links the latent transformation with the text descriptions under the guidance of CLIP~\cite{radford2021learning}.
However, the discovered editing directions still need to be manually labeled for meaningful editing, and it is hard to produce the user-desired directions.

In this paper, we study an adaptive nonlinear transformation rather than linear interpolation in~\cite{shen2020interpreting,yao2021latent,khrulkov2021latent,harkonen2020ganspace,shen2021closed,yang2021discovering,abdal2022clip2stylegan}, and investigate a density regularization to encourage in-distribution latent transformation.
\section{The Proposed \methodname}
In this section, we first formulate the problem of face editing and present our motivation in Sec.~\ref{sec:motivation},
then describe the proposed adaptive nonlinear transformation in Sec.~\ref{sec:ada_trans} and latent density regularization in Sec.~\ref{sec:latent_reg},
followed by the training and inference of the proposed \methodname in Sec.~\ref{sec:training}.
Fig.~\ref{fig:framework} illustrates the framework of the proposed \methodname.

\subsection{Problem Formulation and Motivation}
\label{sec:motivation}

As presented in Fig.~\ref{fig:framework},
we are now given a classifer network $Q$ to estimate the facial attributes $\mat{a}=\{a_1, a_2, \ldots, a_N\}$, where $N$ is the number of attribute annotations, and $a_i$ can be binary or one-hot value.
Face editing focuses on manipulating one or multiple attributes in $\mat{a}$ without changing others.
Another pre-trained StyleGAN generator $G$ is employed to produce high-resolution photorealistic faces $\mathcal{I}\subset \mathbb{R}^{H \times W}$ from the latent space $\mathcal{W}\subset \mathbb{R}^{512}$, where $H\times W$ represents the image size.

To perform face editing with $G$, the face $\mat{I}\in \mathcal{I}$ should be inverted into the latent space with an encoder $E$: $\mathcal{I}\to \mathcal{W}$ to obtain the latent code $\mat{w}=E(\mat{I})$.
The practical choices~\cite{tov2021designing,alaluf2021restyle,richardson2021encoding} tend to adopt $\mathcal{W}^+\subset \mathbb{R}^{18\times 512}$ with layer-wise latent codes for faithful reconstruction.
Then, the inverted latent codes are manipulated to change the target attributes. The resulting edited latents $\mat{w}_\mathrm{e}$ can be fed into $G$ to obtain the generated face $G(\mat{w}_\mathrm{e})$.

Previous literature~\cite{shen2020interpreting,yao2021latent,harkonen2020ganspace,shen2021closed,abdal2022clip2stylegan} usually edits the facial attribute $a_i$ by linear interpolation in the latent space with certain editing directions $\vct{n}_{a_i}\in \mathbb{R}^{512}$, which can be formulated as follows:
\begin{align}
    \mat{w}_{\mathrm{e}} = \mat{w} + \alpha \vct{n}_{a_i}.
    \label{eq:linear}
\end{align}
In such a setting, $\alpha\in \mathbb{R}$ is a scalar to control the editing strength of manipulating the face to $a_i$. $\vct{n}_{a_i}$ can be learned by training a hyperplane/fully-connected layers in the latent space~\cite{shen2020interpreting,yao2021latent}, or discovered from the principal components~\cite{harkonen2020ganspace,shen2021closed} and text information~\cite{abdal2022clip2stylegan}.

\begin{figure}[t]
    \centering
    \includegraphics[width=0.925\linewidth]{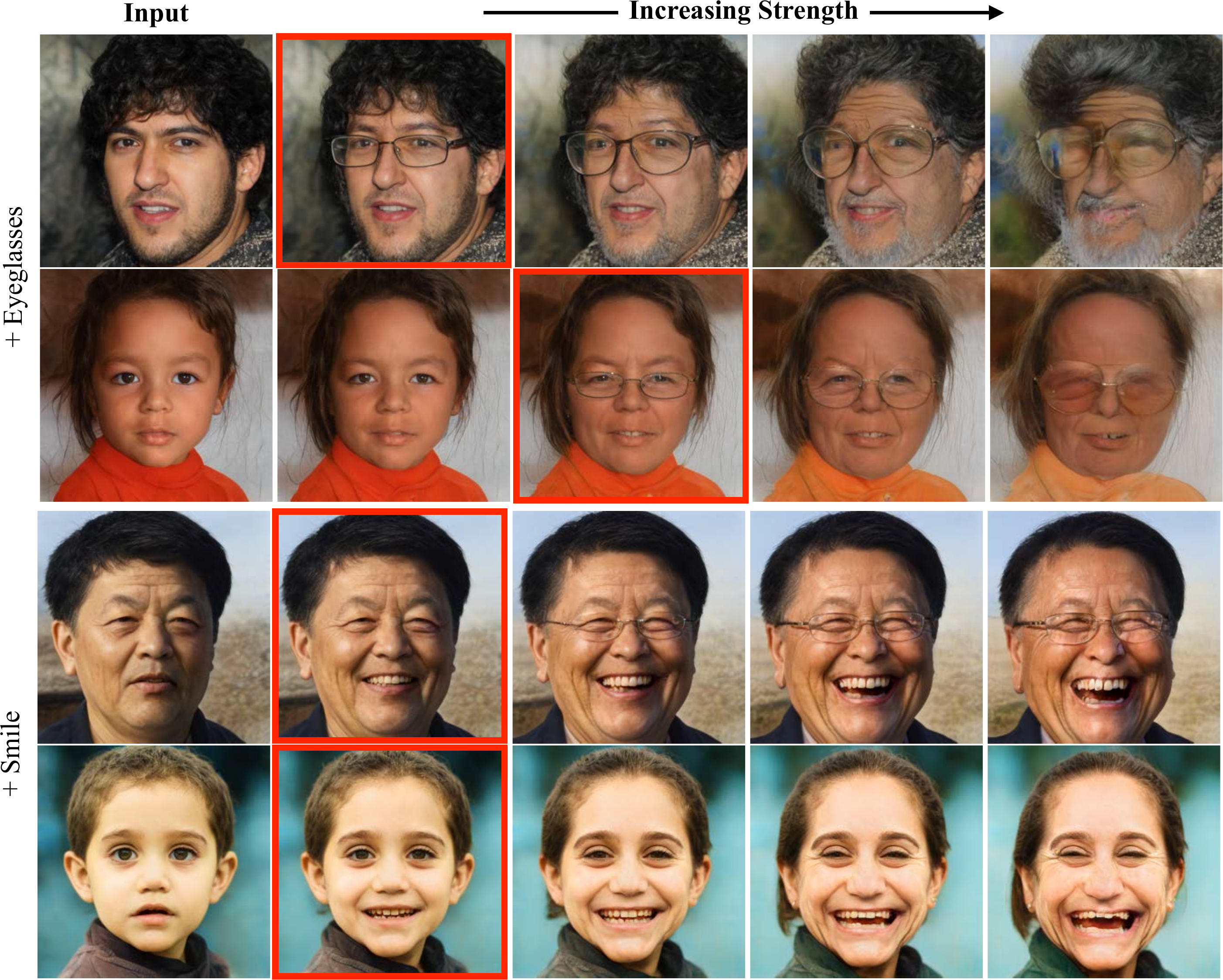}
    \caption{
    Sample results of linear interpolation of InterFaceGAN~\cite{shen2020interpreting} varying the strength. The leftmost text describes the changed attributes, and the red boxes indicate that the faces have been successfully manipulated to the target attributes. With the increase of strength, the unrelated attributes are changed and fidelity is harmed.
    \label{fig:illustration1}
    }
\end{figure}

However, as observed in Fig.~\ref{fig:illustration1}, there are several limitations in such simple linear interpolation. \emph{First}, linearly manipulating the latent code would change other unrelated attributes due to the entanglement of the latent space. \emph{Second}, the optimal strength for accurate editing is hard to tune. Small strength may not change the desired attributes while a large one would harm the face quality as the latent codes are out of the latent space. Importantly, the strength varies from input faces.
\emph{Last}, it is limited to binary attributes such as \texttt{gender} and \texttt{young}, and cannot handle those more complicated attributes, \eg, finer age that cannot be controlled by a linear transformation. To verify this claim, supplementary Fig.~\ref{fig:scatter_point_age} visualizes the latent codes with fine age labels. 

Consequently, we need an adaptive transformation for nonlinear editing to address the above issues, which will be detailed in the following sections.

\subsection{Adaptive Nonlinear Transformer}
\label{sec:ada_trans}

To avoid tuning the optimal strength for face editing with linear interpolation, we propose an adaptive nonlinear transformer for traversing the latent space of GANs. 
Specifically, instead of directly learning an editing direction, we opt to divide the whole transformation process into several fine steps, where the size and direction at each step are conditioned on the target attributes. 
As a result, a nonlinear transformation trajectory can be obtained by these fine linear steps and the endpoint is the edited latent code that we are desired.

As shown in Fig.~\ref{fig:framework}, our adaptive transformer takes the latent code $\mat{w}$ as the input. An LSTM~\cite{graves2012long} is also employed at the beginning to smooth the transformation trajectory.
At step $t$, learnable affine transformations are adopted to inject the changed/unchanged attributes to modulate the size and the direction of manipulation, which are parameterized by a two-layer multilayer perceptron (MLP) following the AdaIN~\cite{huang2017arbitrary} operation:
\begin{align}
\mathrm{AdaIN}(\mat{h}^{(t)}_j, \mat{a}) = \mat{y}^{(t)}_{j,s}(\mat{a}) \mat{h}^{(t)}_j + \mat{y}^{(t)}_{j,b}(\mat{a}),
\end{align}
where $\mat{h}_j$ is the input feature, and $\mat{y}_{j,s}(\cdot)$ and $\mat{y}_{j,b}(\cdot)$ output the learned scale and bias, respectively.

The intermediate edited latent code at $t$ step is formed as:
\begin{align}
    \mat{w}^{(t)}_{\mathrm{e}} = \mat{w}^{(t-1)}_{\mathrm{e}} + s^{(t)} \vct{n}^{(t)},
\end{align}
where $\mat{w}_{\mathrm{e}}^{(0)}=\mat{w}$, and the adaptive transformer $f$ outputs $\{\vct{n}_t, s_t\}=f(\mat{w}^{(t-1)}_{\mathrm{e}}, \mat{a})$ conditioned on the target attributes $\mat{a}$. Here, $s_t\in (0, 1)$ activated by a sigmoid function is a scalar to control the step size, and $\vct{n}$ is a unit vector, \ie, $\|\vct{n}\|_2=1$.
Consequently, our adaptive transformer can adaptively adjust the latent code in terms of previously manipulated results.

In summary, the manipulation process can be described as the combination of the intermediate states as follows:
\begin{align}
    \mat{w}_{\mathrm{e}} = \mat{w} + \sum_{t=1}^M s^{(t)} \vct{n}^{(t)},
\end{align}
where $M$ is the maximum steps of the trajectory.

\subsection{Latent Density Regularization}
\label{sec:latent_reg}

Although benefitting from the well-trained StyleGAN generator, the latent space of StyleGAN generator is fixed during editing and hence over-manipulating the latent codes to the target attributes would harm the image quality since the codes are out of the distribution $\mathcal{W}$. It is natural to observe this problem. 
For example, in Fig.~\ref{fig:illustration1}, when the strength of manipulation is set to a large value, the faces have been changed a lot to target attributes at the cost of unnatural generated faces, since they are far away from decision boundaries~\cite{shen2020interpreting}.

The simple solution is to employ an additional loss to restrict the Euclidean distance between the original and edited latent code~\cite{yao2021latent}, which can be written as:
\begin{align}
\mathcal{L}_{\mathrm{dist}}=\|\mat{w}_{\mathrm{e}}-\mat{w}\|_2.
\label{eq:recon_loss}
\end{align}
However, it cannot deal with the essential nature of this problem with the following problems: (1) simply restricting the editing strength may fail to change the desired attributes, and (2) this problem still remains when increasing the strength. Another solution is to train a discriminator again to improve face quality, which, however, will significantly increase the computational cost and the training difficulty.

\begin{figure*}[t]
    \centering
    \includegraphics[width=1.0\linewidth]{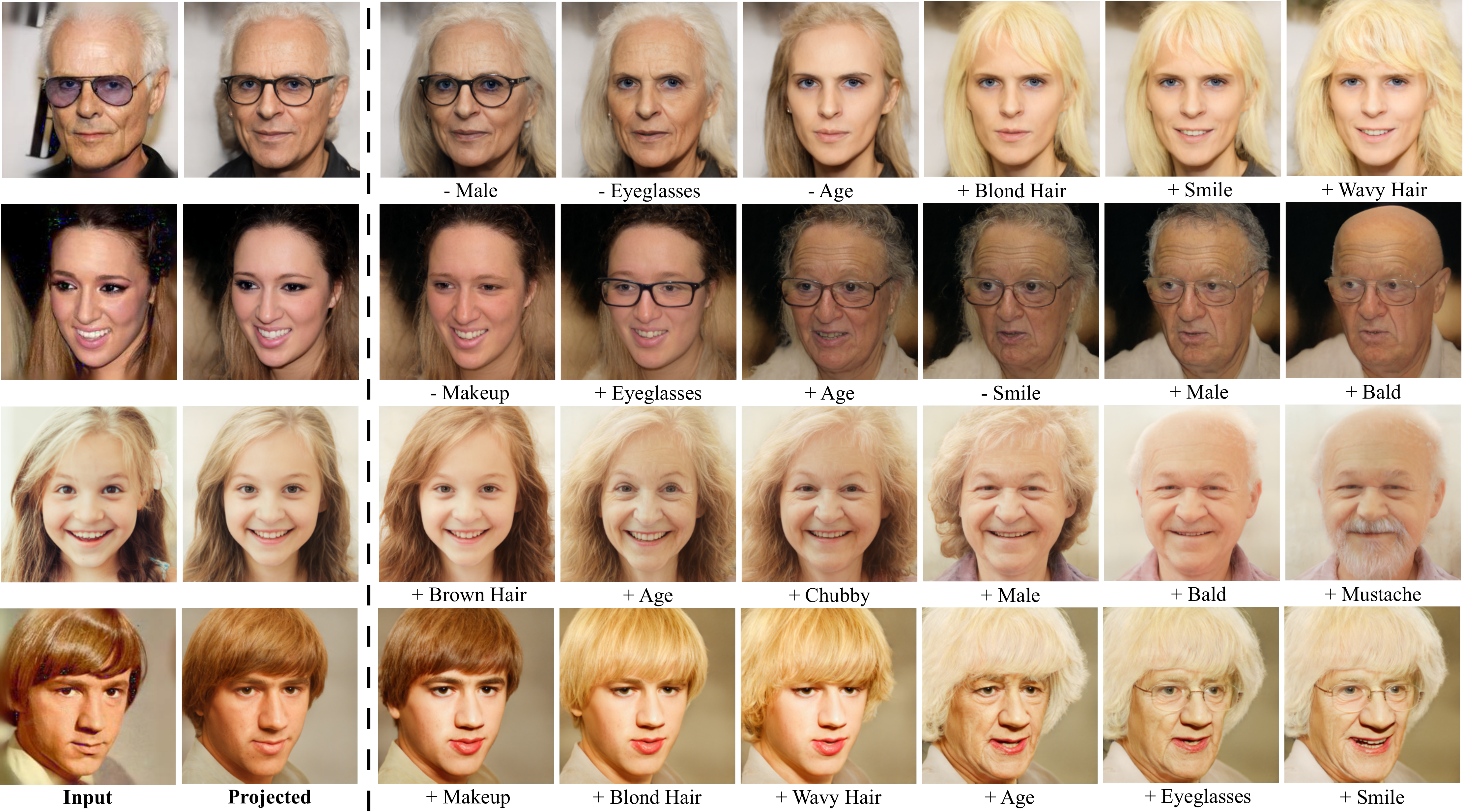}
    \vspace{-20pt}
    \caption{
        Disentangled and controllable face editing on real faces with binary facial attributes. Given an input face, we gradually manipulate it into the target attributes underneath, while keeping the previous manipulating effects.
    }
    \label{fig:qualitative_results2}
\end{figure*}

To address this issue, we propose to use a regularization term to encourage the edited latent codes to fall into the distribution of the latent space. We employ a simple yet effective method by maximizing the likelihood of the transformation trajectory with a pre-trained density model. Here, the density model is trained in advance to model the latent space. In this paper, we adopt the real NVP~\cite{dinh2016density} due to its simplicity and effectiveness in modeling data distribution.

In doing so, we can always restrict the transformation trajectory in the latent space, so that the fidelity of generated faces would not be distorted. The resultant regularization term can be thus written as:
\begin{align}
\mathcal{L}_{\mathrm{reg}}=-\frac{1}{M} \sum_{t=1}^M \log p_\phi(\mat{w}^{(t)}_{\mathrm{e}}).
\label{eq:nll_loss}
\end{align}
Here, the density model is parameterized by $\phi$ and $M$ is the maximum step of the trajectory.

\subsection{Training and Inference}
\label{sec:training}

\paragraph{Disentangled learning.} Given the attributes $\mat{a}=\{a_1, \hat{a}_2, \ldots, a_N\}$ to be changed or kept, we propose to achieve disentangled face editing by maximizing the mutual information $\mathbb{I}(\mat{a}; G(\mat{w}_{\mathrm{e}}))$ between conditions and generated faces, following~\cite{chen2016infogan}. Formally, the mutual information loss function can be written as:
\begin{align}
    \mathcal{L}_{\mathrm{mi}}=-\sum_{a\in \mat{a}}\log p_\theta(a|G(\mat{w}_{\mathrm{e}})),
\label{eq:mi_loss}
\end{align}
where the true posterior $p_\theta(a|G(\mat{w}_{\mathrm{e}}))$ is approximated by the classifier $Q$. In the supervised setting, $Q$ is pre-trained on the labeled dataset and fixed during training.
We would like to highlight that \methodname puts all attributes in $\mat{a}$, which enables \methodname to attenuate the entanglement between attributes~\cite{chen2016infogan}.
The most related work to \methodname is~\cite{abdal2021styleflow} which models the conditional distributions of latent codes and attributes using continuous normalizing flows~\cite{chen2018neural}. However, the sampling procedure in~\cite{abdal2021styleflow} makes it difficult to handle the scenario with few labeled examples, as studied in Sec.~\ref{sec:exp}.

\paragraph{Training and inference.}
Combining Eqs.~\eqref{eq:recon_loss}, \eqref{eq:nll_loss}, and  \eqref{eq:mi_loss}, the overall training objective for \methodname is to minimize the sum of all losses:
\begin{align}
    \mathcal{L}=
    \lambda_{\mathrm{dist}} \mathcal{L}_{\mathrm{dist}} +
    \lambda_{\mathrm{reg}} \mathcal{L}_{\mathrm{reg}} + 
    \lambda_{\mathrm{mi}} \mathcal{L}_{\mathrm{mi}},
\end{align}
where $\lambda_{\mathrm{*}}$ controls the importance between different loss components. Here, $\mathcal{L}_{\mathrm{reg}}$ is employed to compress those unnecessary changes in the latent codes. During training, the binary attributes of the given face are randomly manipulated to $\{0, 1\}$. As a result, \methodname can manipulate the latent code to desired attributes while keeping others unchanged.
\section{Experiments}
\label{sec:exp}

\begin{figure*}[t]
	\centering
	\subfloat[Editing single attributes.]{
        \label{fig:quantitative_results_single_attributes}
		\includegraphics[width=.48\columnwidth]{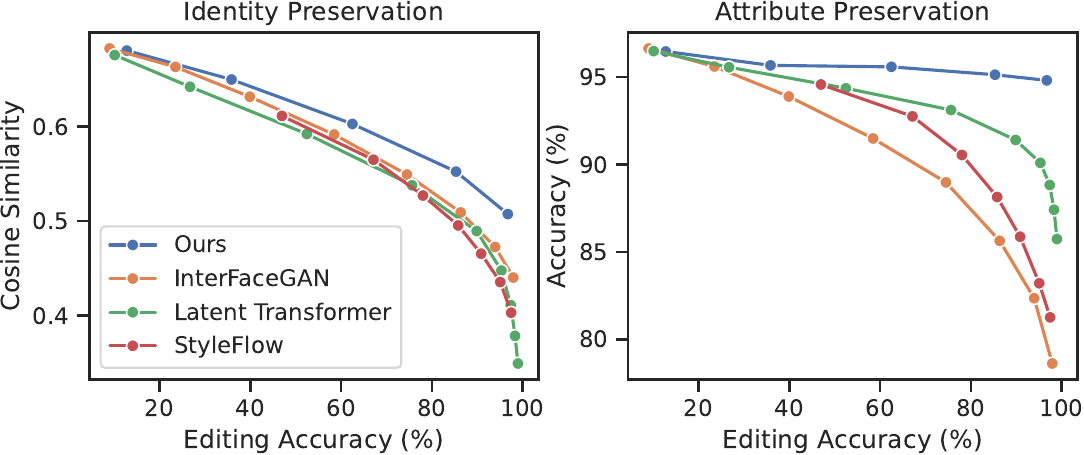}
		}
		\vspace{-10pt}
	\subfloat[Editing multiple attributes.]{
        \label{fig:quantitative_results_multi_attributes}
		\includegraphics[width=.48\columnwidth]{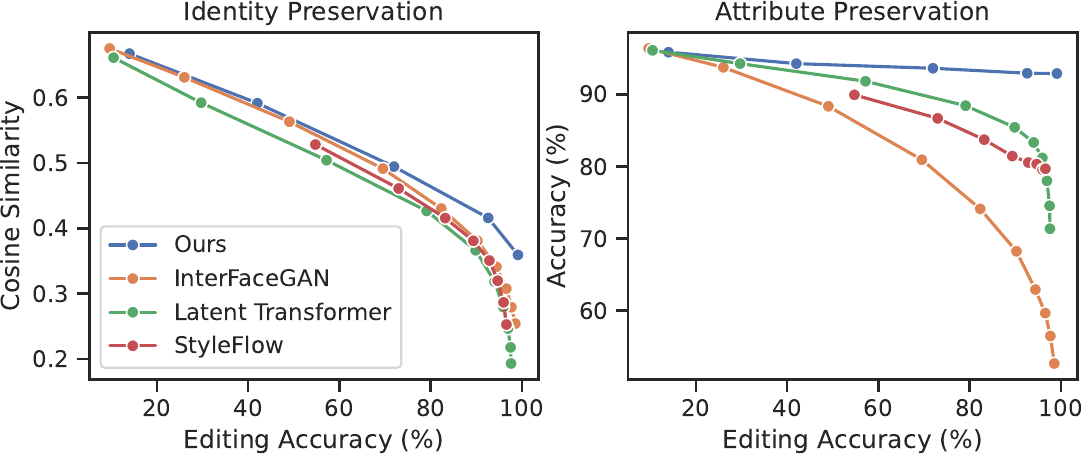}
		}\\
	\subfloat[Editing with large age gap.]{
        \label{fig:quantitative_results_age_gap}
		\includegraphics[width=.48\columnwidth]{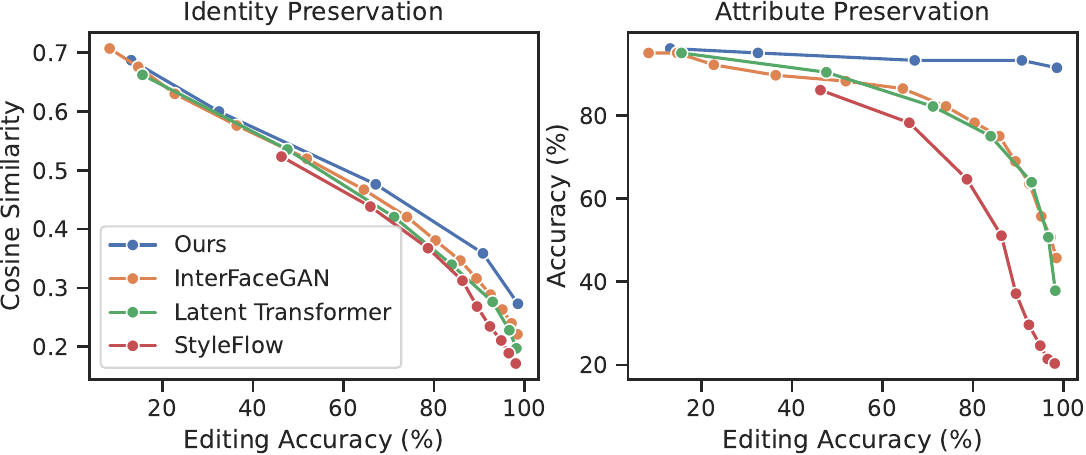}
		}
	\subfloat[Editing with limited labeled data.]{
        \label{fig:quantitative_results_limited_data_single_attributes}
		\includegraphics[width=.48\columnwidth]{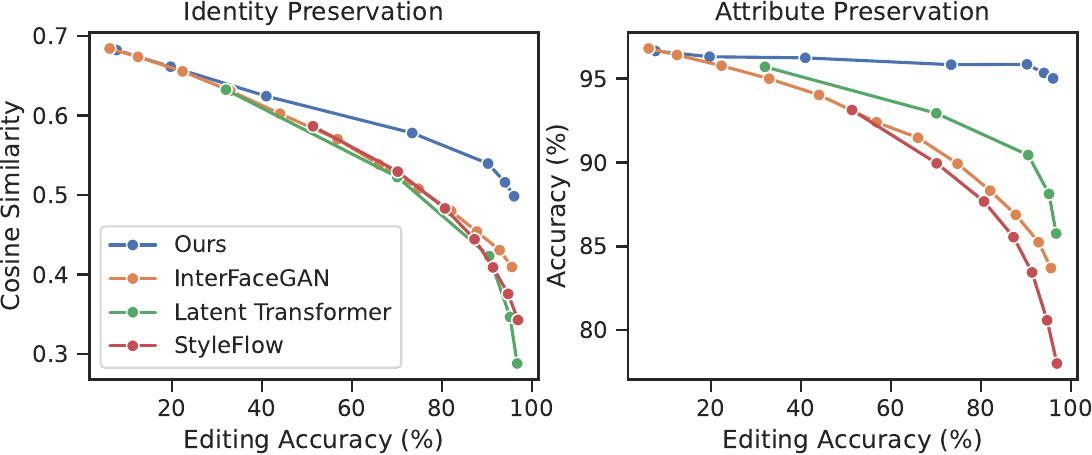}
		}
	\caption{Quantitative comparisons with recent methods for face editing under different experimental settings. We employ attribute/identity preservation during editing to evaluate different methods. We desire higher editing accuracy while keeping the original identity and unrelated attributes. Therefore, a higher curve indicates better performance.
    }
        \label{fig:quantitative_comparisons}

\end{figure*}

\subsection{Implementation Details}
In this paper, we perform face editing in the latent space of the StyleGAN2 generator~\cite{karras2020analyzing} pre-trained on FFHQ~\cite{karras2019style} and employ e4e encoder~\cite{tov2021designing} to project the faces into $\mathcal{W}^+$ space. For the fixed attribute classifier, we trained the last linear layer of ResNet-50~\cite{he2016deep} on the binary attributes of CelebA dataset~\cite{liu2015deep} and the discrete age labels of FFHQ from~\cite{or2020lifespan}. We train another classifier from scratch for better classification performance and obtain the attribute labels of FFHQ dataset.
The first 69k faces of FFHQ are left as training data with an image size of $256\times 256$. The rest 1k faces and CelebA-HQ~\cite{karras2017progressive} are used as the testing data.

\methodname is trained for 10k iterations with a batch size of 16, Adam optimizer~\cite{kingma2014adam} with a fixed learning rate of $10^{-4}$, $\beta_1=0.9$ and $\beta_2=0.99$. A one-layer LSTM and 10 two-layer MLP with the dimension 512 and ReLU activation are stacked. The direction and size of each step are output by a separate full-connected layer. The loss weights $\lambda_{\mathrm{*}}$ and the maximum step $M$ are set to 1 and 5, respectively. Real NVP~\cite{dinh2016density} is employed as the density model trained on the latent codes without any supervision for 10k iterations.

\subsection{Disentangled and Controllable Face Editing}

\subsubsection{Qualitative Results} 

Fig.~\ref{fig:qualitative_results2} showcases example results on manipulating the faces into target attributes. We project the faces into the $\mathcal{W}^+$ space of StyleGAN2 using~\cite{tov2021designing}, and then gradually manipulate the latent codes with a sequential attribute list. Obviously, \methodname achieves photorealistic and disentangled modifications on the resultant faces, \ie, the identity and other attributes are well preserved during the manipulation process. Note that we train all attributes presented in Fig.~\ref{fig:qualitative_results2} in a unified model. Therefore, \methodname has achieved great flexibility and controllability simultaneously and is not limited to a defined order of attributes. 
We provide additional examples for sequential editing in supplementary Fig.~\ref{fig:supp-sequence} and the results for editing a single attribute in supplementary Fig.~\ref{fig:supp_single_attribute_editing}.

\subsubsection{Comparisons with State-of-the-arts} 

\paragraph{Competitors.} 
To validate the effectiveness of \methodname, we perform comparisons with the recent state-of-art methods including InterFaceGAN~\cite{shen2020interpreting}, StyleFlow~\cite{abdal2021styleflow} and Latent Transformer~\cite{yao2021latent}. We implement all methods on the latent codes of FFHQ dataset projected by~\cite{tov2021designing}, strictly following their experimental settings. We highlight that editing on real faces is much more difficult than synthetic ones. Only three binary attributes for \texttt{gender}, \texttt{eyeglasses}, and \texttt{young} are adopted to evaluate different methods, which are definitely shown to be entangled together~\cite{shen2020interpreting} with each other in the latent space of StyleGAN. For fair comparisons, the same attribute classifier is used for Latent Transformer.

\begin{figure*}[t]
	\centering
	\subfloat[Editing single attributes.]{
        \label{fig:qualitative_comparisons_single_attributes}
		\includegraphics[width=.47\textwidth]{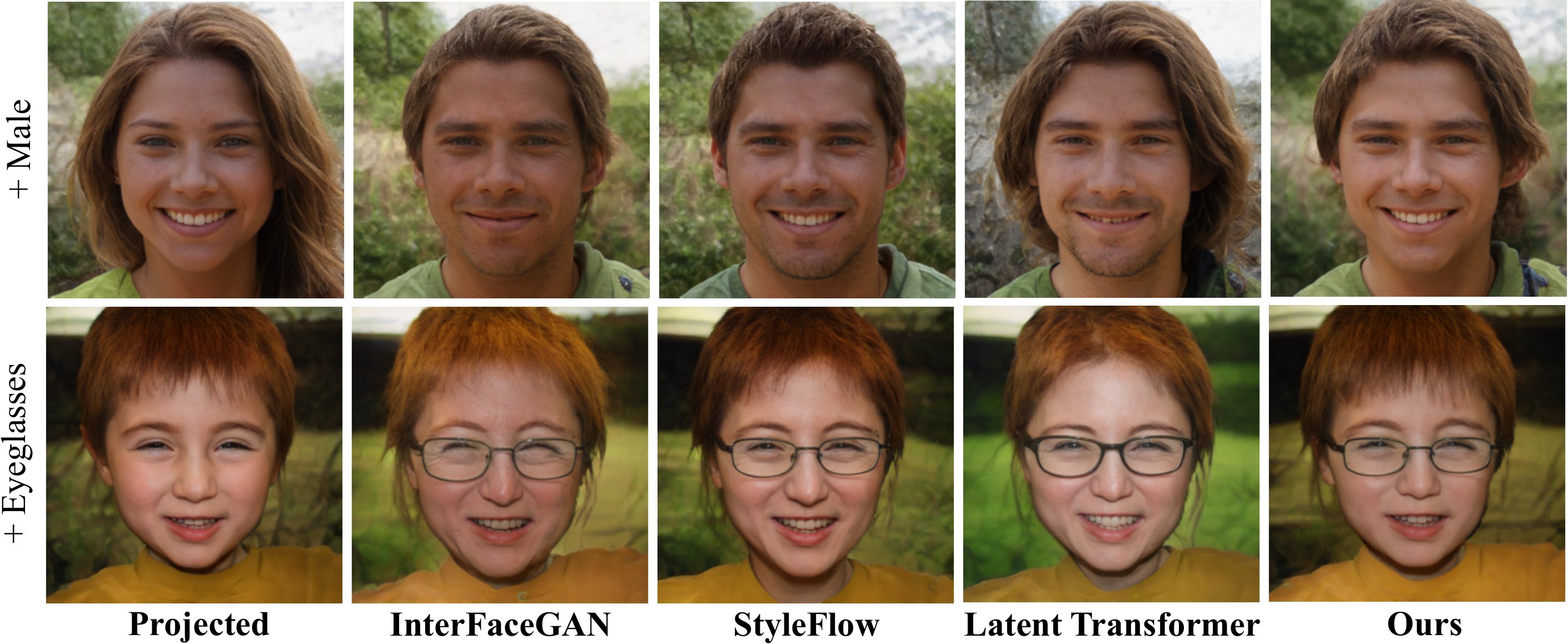}
		}
		\vspace{-10pt}
	\subfloat[Editing multiple attributes.]{
        \label{fig:qualitative_comparisons_multi_attributes}
		\includegraphics[width=.47\textwidth]{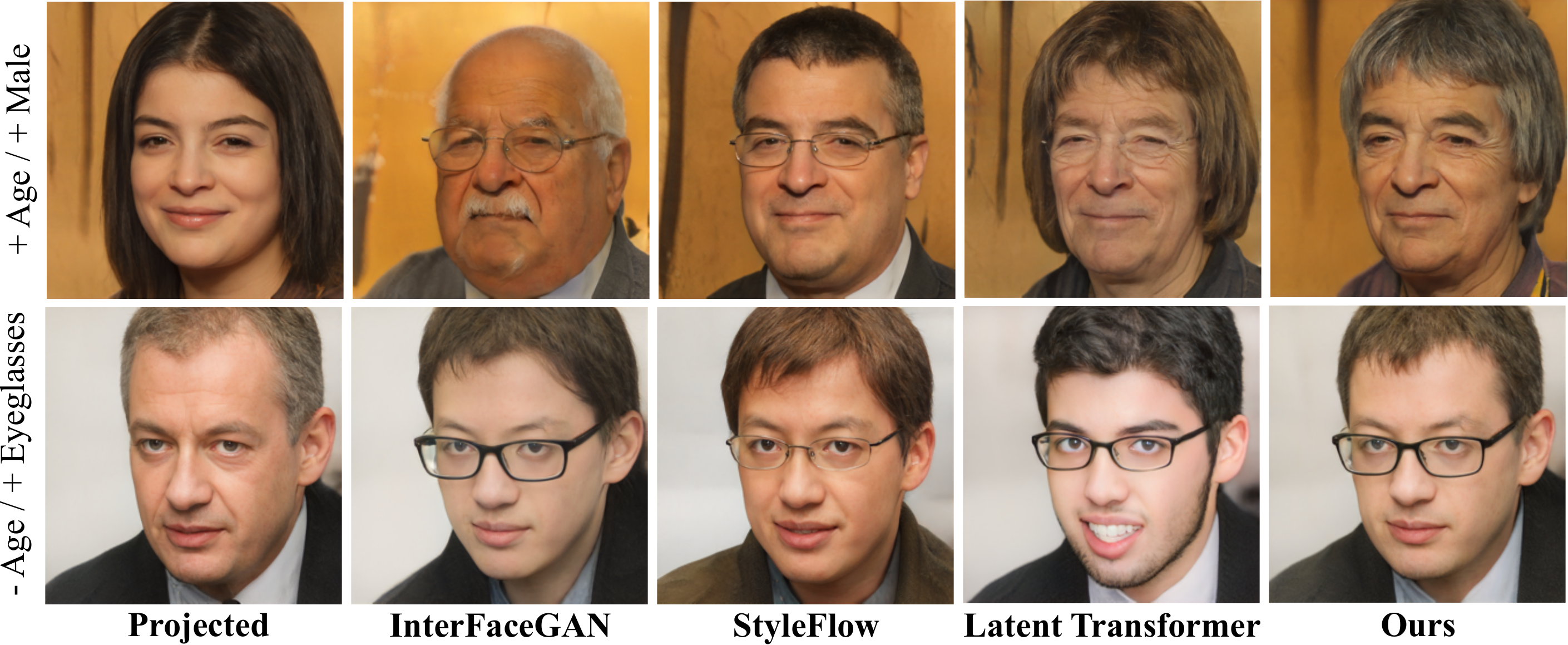}
		}\\
	\subfloat[Editing with large age gap.]{
        \label{fig:qualitative_comparisons_age_gap}
		\includegraphics[width=.47\textwidth]{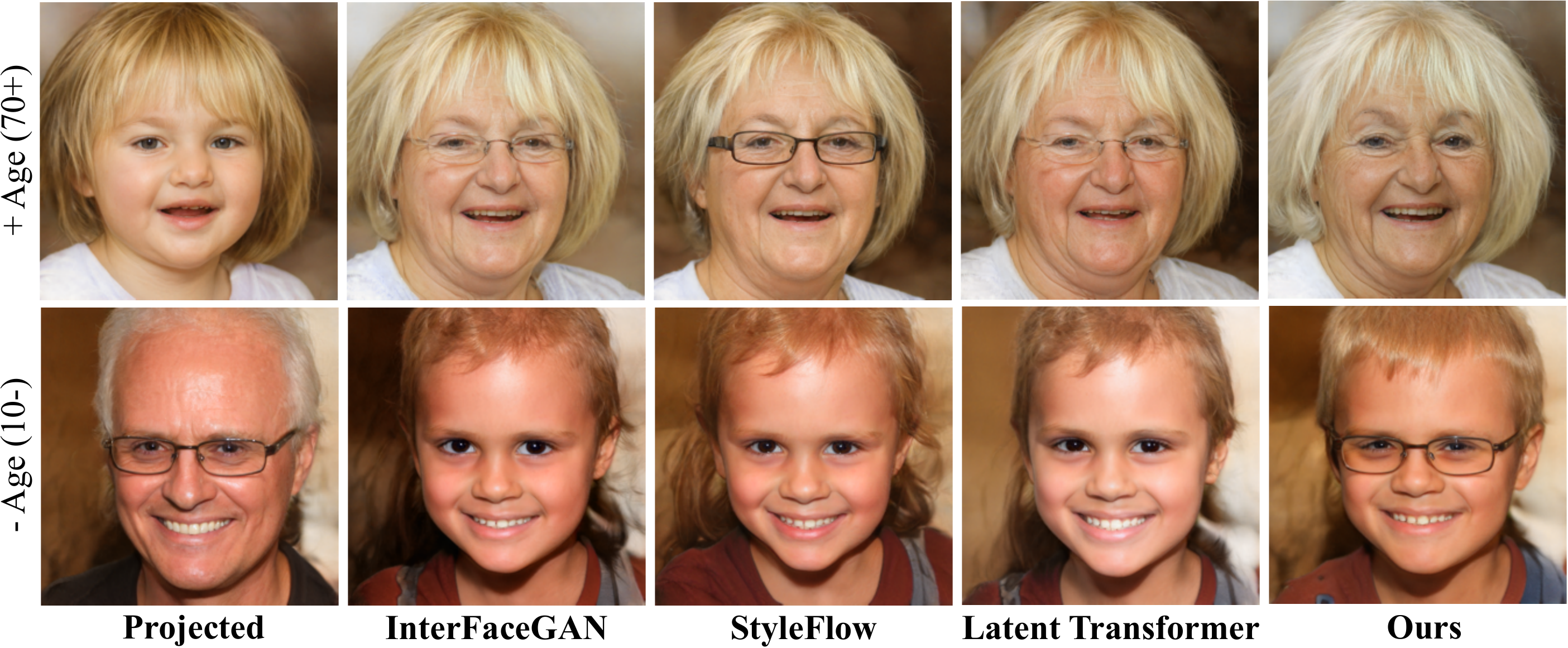}
		}
	\subfloat[Editing with limited labeled data.]{
        \label{fig:qualitative_comparisons_limited_data_single_attributes}
		\includegraphics[width=.47\textwidth]{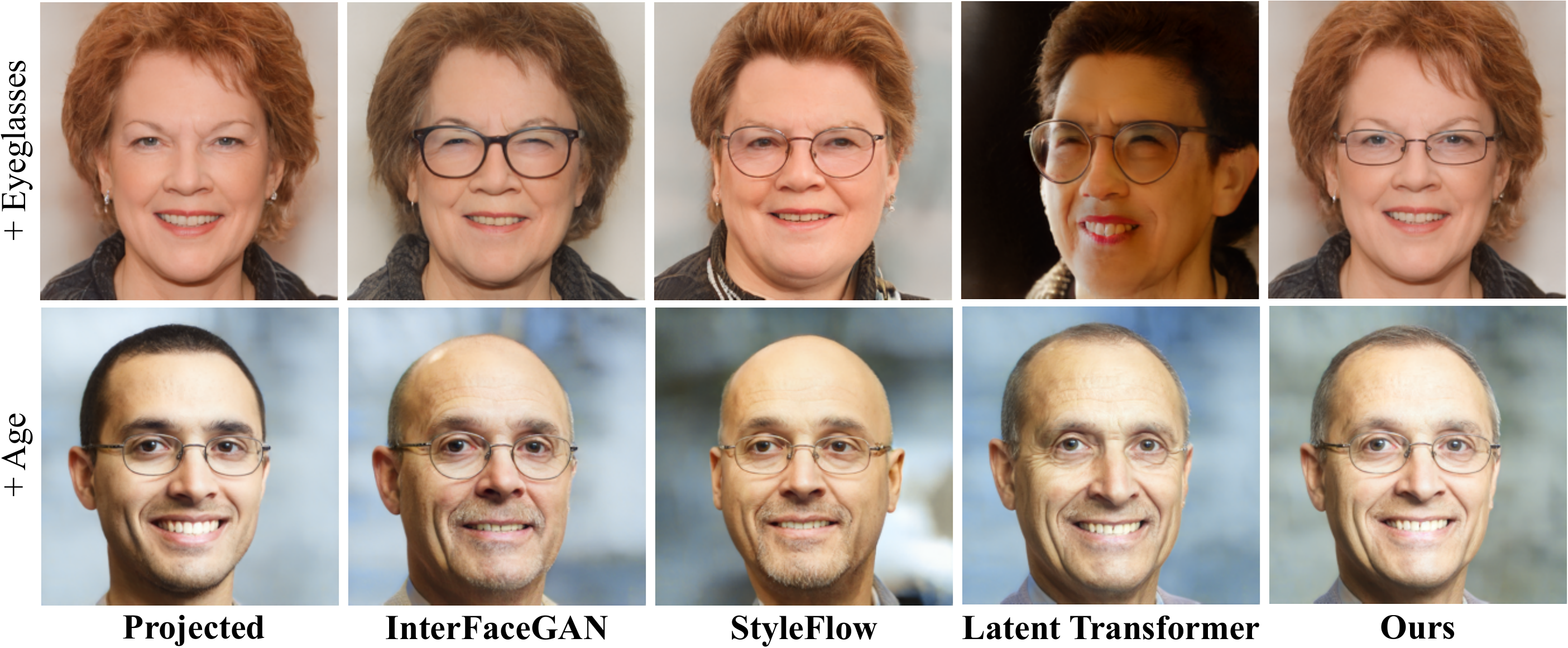}
		}
		\vspace{-2.0pt}
	\caption{Qualitative comparisons with recent methods for face editing under different experimental settings.
	The editing strengths of different methods are manually increased until the faces are manipulated into target attributes. The competitors produce unexpected changes in unrelated attributes or fail to handle the scenarios with extreme conditions like age gaps and few labeled samples.
    }
      \label{fig:qualitative_comparisons}

\end{figure*}

\paragraph{Evaluation metrics.} 
We employ three widely-used metrics to compare different methods quantitatively, including editing accuracy, attribute preservation accuracy, and identity preservation. A ResNet-50 is trained from scratch to predict the attributes of manipulated faces, which can be used to measure the editing accuracy and attribute preservation for different facial attributes. The editing accuracy indicates the proportion of the samples that have been successfully manipulated into target attributes, \ie, the probability is greater than $0.5$. On the other hand, attribute preservation accuracy is the proportion of the rest attributes being kept. Identity preservation is the cosine similarity between the facial embeddings of the original input and manipulated faces extracted by ~\cite{deng2019arcface}. In a sense, we desire a higher editing accuracy~(\ie, manipulating the faces into target attributes) with fewer drops in identity and attribute preservation~(\ie, identity and other unrelated attributes are unchanged).

The input faces in testing data are manipulated into the opposite attribute with different strengths to control the degree of manipulation.
We gradually increase the strengths for different methods as suggested by their official code until the editing accuracy reaches 99\%. More specifically, we interpolate the label variables of StyleFlow~\cite{abdal2021styleflow}, scale the learned directions for InterFaceGAN~\cite{shen2020interpreting} and Latent Transformer~\cite{yao2021latent}, and modifies the maximum steps $M$ of \methodname. Consequently, we can draw the curves of identity/attribute preservation w.r.t. editing accuracy. 

Figs.~\ref{fig:quantitative_comparisons} and~\ref{fig:qualitative_comparisons} show the comparisons between different methods under different settings. Additional qualitative comparisons are provided in supplementary Fig.~\ref{fig:supp_qualitative_comparisons}.

\paragraph{Editing single/multiple attributes.} 
In this setting, each single or any two~(\ie, multiple) attributes are manipulated at the same time.
The quantitative results are shown in Figs.~\ref{fig:quantitative_results_single_attributes} and~\ref{fig:quantitative_results_multi_attributes}.
Obviously, \methodname outperforms the other three competitors by a large margin. We showcase the qualitative results in Figs.~\ref{fig:qualitative_comparisons_single_attributes} and~\ref{fig:qualitative_comparisons_multi_attributes}. For attributes like becoming old, the glasses appear for other methods, indicating that the entangled attributes are not handled well. Overall, the results demonstrate that \methodname achieves better disentangled and accurate face editing both quantitatively and qualitatively. 

\paragraph{Editing with large age gaps.}
As studied in the previous setting, the most challenging task is becoming old as the glasses and age are entangled together in the latent space. We further validate the effectiveness of \methodname in a much more rigorous setting, \ie, manipulating the children below 10 years old to the old above 70 years old. We replace the \texttt{young} label with the age labels of FFHQ dataset provided by~\cite{or2020lifespan}, which consists of 7 discrete age groups. The quantitative results in Fig.~\ref{fig:quantitative_results_age_gap} show that \methodname performs significantly better that the competitors. We note that it is natural to see a severe drop in identity preservation since identity would be changed a lot during the facial aging process. In terms of qualitative results in Fig.~\ref{fig:qualitative_comparisons_age_gap}, \methodname can still preserve the eyeglasses from 70+ to 10-, and eliminate the appearance of eyeglasses from 10- to 70+. In summary, in the most challenging experimental settings, \methodname can still achieve disentangled face editing over state-of-the-art methods.

\paragraph{Editing with limited labeled data.}
An important problem of supervised face editing methods~\cite{abdal2021styleflow,shen2020interpreting,yao2021latent} is that they require a large number of labeled samples, which is difficult to collect in practical scenarios. In this experimental setting, only 128 labeled samples for each attribute are used to challenge different editing methods. The quantitative and qualitative results for single attribute manipulation are shown in Fig.~\ref{fig:quantitative_results_limited_data_single_attributes} and Fig.~\ref{fig:qualitative_comparisons_limited_data_single_attributes}, respectively. The proposed \methodname can still handle well and achieve photorealistic face editing with limited labeled samples, benefitting from the disentangled learning strategy. Therefore, \methodname is more flexible to practical applications than the competitors.
An ablation study on the number of labeled samples is conducted in supplementary section~\ref{sec:ablation_labeled_data}.

\subsection{Ablation Study}

We conduct ablation studies to validate the effectiveness of different components in \methodname.

\begin{figure}[t]
    \centering
    \includegraphics[width=1.0\linewidth]{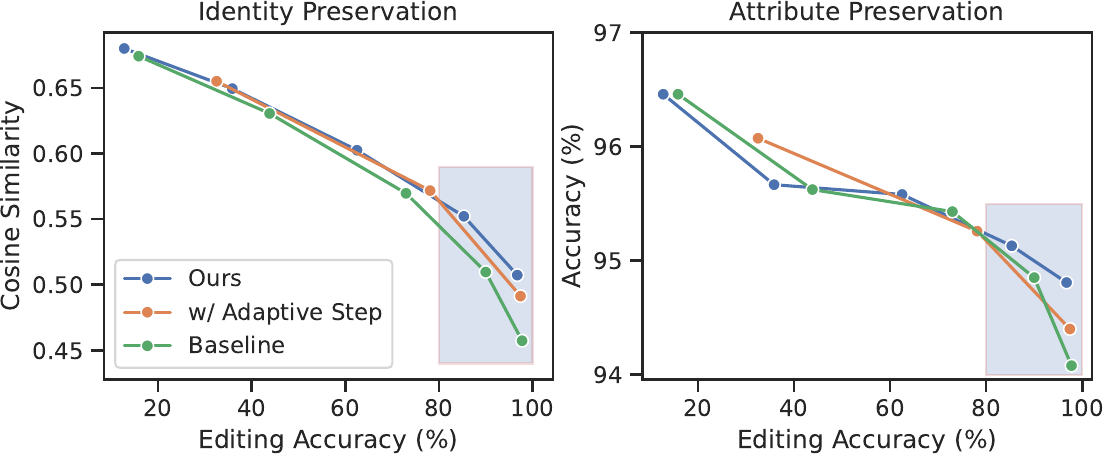}
    \caption{
        Quantitative results of the proposed components.
		 The color box is the region of interest for comparisons.
    }
    \label{fig:ablation_quantitative}
\end{figure}

\paragraph{Ablation study on the proposed components.}
We start from the baseline method without the proposed adaptive nonlinear transformer and density regularization, directly optimized by $\mathcal{L}_{\mathrm{mi}}$ and $\mathcal{L}_{\mathrm{dist}}$. The baseline still performs nonlinear editing, however, with a fixed step size at each step, compared to the full \methodname. Then, we gradually apply the proposed adaptive strategy and latent density regularization to the baseline, leading to two variants of our \methodname.

In Fig.~\ref{fig:ablation_quantitative}, significant drops in identity/attribute preservation have arisen for the baseline; see the color box. We argue that the baseline cannot compress the unnecessary changes in the latent codes and does not adjust the manipulation process with the fixed step size.

On the other hand, the adaptive strategy can address this problem by adaptively controlling the manipulation according to the previous trajectory and target attributes. Furthermore, we have also found the fidelity of generated faces has been unexpectedly harmed. We show the qualitative results in supplementary Fig.~\ref{fig:ablation_qualitative}.
To be manipulated into ideal target attributes, the latent codes may fall out of the latent space, which, as a result, harms the image quality produced by a pre-trained StyleGAN generator. This phenomenon motivates us to employ latent regularization to encourage the edited latent codes to be in-distribution.

\begin{figure}[t]
    \centering
    \includegraphics[width=1.0\linewidth]{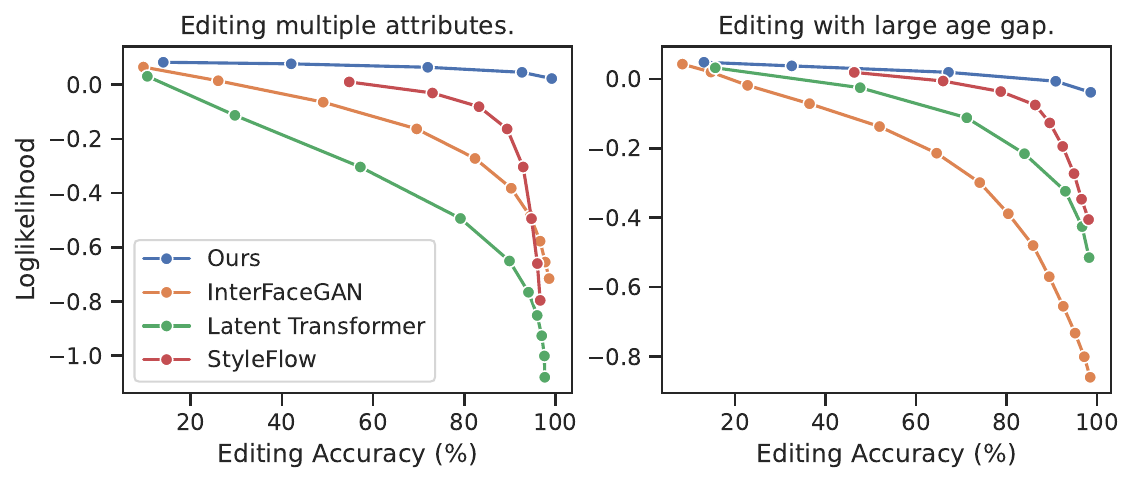}
    \caption{
        Quantitative comparisons of the loglikelihood.
    }
    \vspace{-5pt}
    \label{fig:loglikelihood}
\end{figure}

\paragraph{Analysis of regularization term.}
The pre-trained density model is employed to estimate the loglikelihood of edited latent codes~(normalized by dimension) for different methods. Fig.~\ref{fig:loglikelihood} presents the results for editing multiple attributes and editing with a large age gap. The higher loglikelihood indicates more in-distribution manipulation, indicating better fidelity of the pre-trained StyleGAN generator. StyleFlow performs better than InterFaceGAN and Latent Transformer as it samples the edited codes from the latent distribution. \methodname is still able to encourage the latent codes in-distribution while achieving better editing accuracy, which is the key idea of latent regularization.

\begin{figure}[h]
    \centering
    \includegraphics[width=1.0\linewidth]{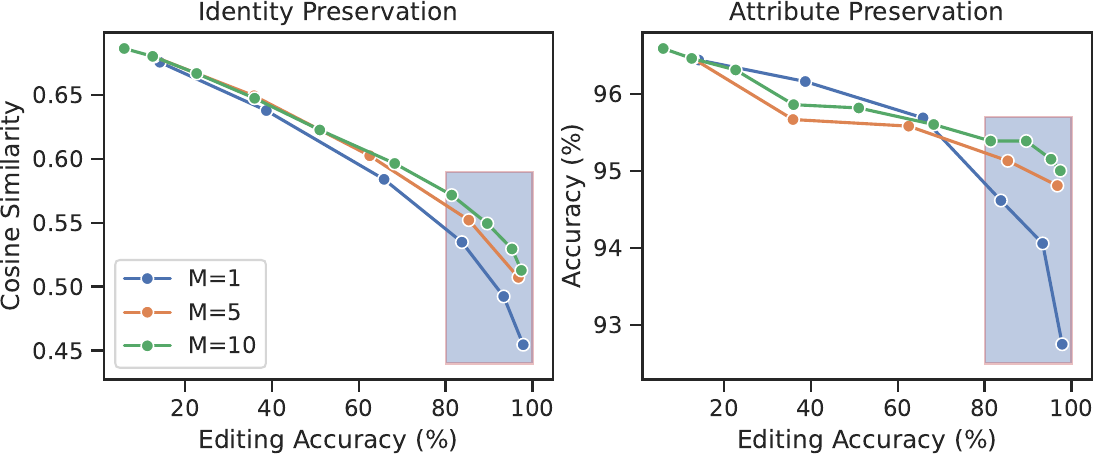}
    \caption{
        Quantitative results for the ablation of $M$.
    }
    \vspace{-5pt}
    \label{fig:ablation_quantitative_M}
\end{figure}

\paragraph{Ablation study on the maximum steps $M$.}
In \methodname, $M$ defines the maximum steps of transformation trajectory. During training, $M$ is empirically fixed and \methodname is trained to learn each step size and direction. During inference, the user can flexibly adjust $M$ for desired visual results~(\eg, becoming younger or older).
Fig.~\ref{fig:ablation_quantitative_M} demonstrates that increasing $M$ during training can further improve editing performance. Although $M$ can be set large enough since \methodname can adaptively adjust the step size, more computational costs may be introduced and effects of each step may be minor. Supplementary Fig.~\ref{fig:supp_different_step} visualizes the transformed trajectory and shows that \methodname has produced smooth transformations.

\section{Conclusion}

In this paper, we introduce \methodname, a novel nonlinear transformation for face editing.
\methodname divides the manipulation process into several finer steps to achieve nonlinear manipulation in the latent space of GANs. 
A predefined density real NVP model regularizes the trajectory of the transformed latent codes to be constrained in the distribution of latent space.
A disentangled learning strategy is employed to eliminate the entanglement among attributes.
Extensive experiments have been conducted to validate the effectiveness of \methodname, especially with the large age gap and few labeled examples.
In the future, we will explore preserving the background for face editing with the help of the intermediate features of StyleGAN generator.

\clearpage
{\small
\bibliographystyle{ieee_fullname}

}

\appendix

\renewcommand \thepart{}
\renewcommand \partname{}
\addcontentsline{toc}{section}{Appendix}
\numberwithin{equation}{section}
\setcounter{figure}{0}
\setcounter{table}{0}
\renewcommand\thetable{A\arabic{table}}
\renewcommand\thefigure{A\arabic{figure}}

\clearpage
\part{\hfill \textsc{Appendix} \hfill}

\section{Ablation study on the number of labeled samples.}
\label{sec:ablation_labeled_data}
A potential advantage of \methodname is the great generalization ability for the scenarios with few limited samples. This advantage benefits from the proposed disentangled learning strategy. To investigate the optimal and least labeled samples used for face editing, experiments are conducted to demonstrate the influence of the number of labeled samples.

\begin{figure}[h]
    \centering
    \includegraphics[width=1.0\linewidth]{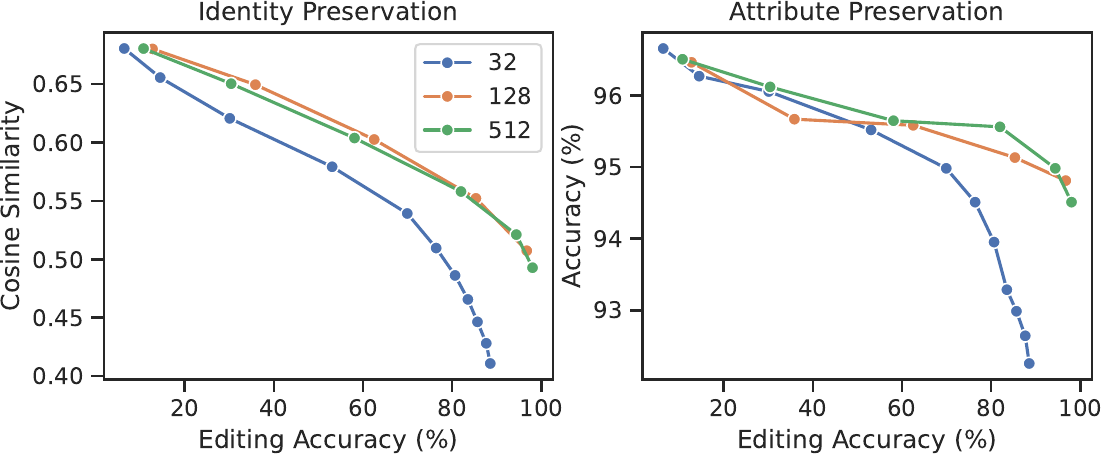}
    \caption{
        Quantitative results for the ablation of the number of labeled samples used for face editing.
    }
    \label{fig:ablation_labeled_data}
\end{figure}

\begin{figure*}[t]
    \centering
    \includegraphics[width=1.0\linewidth]{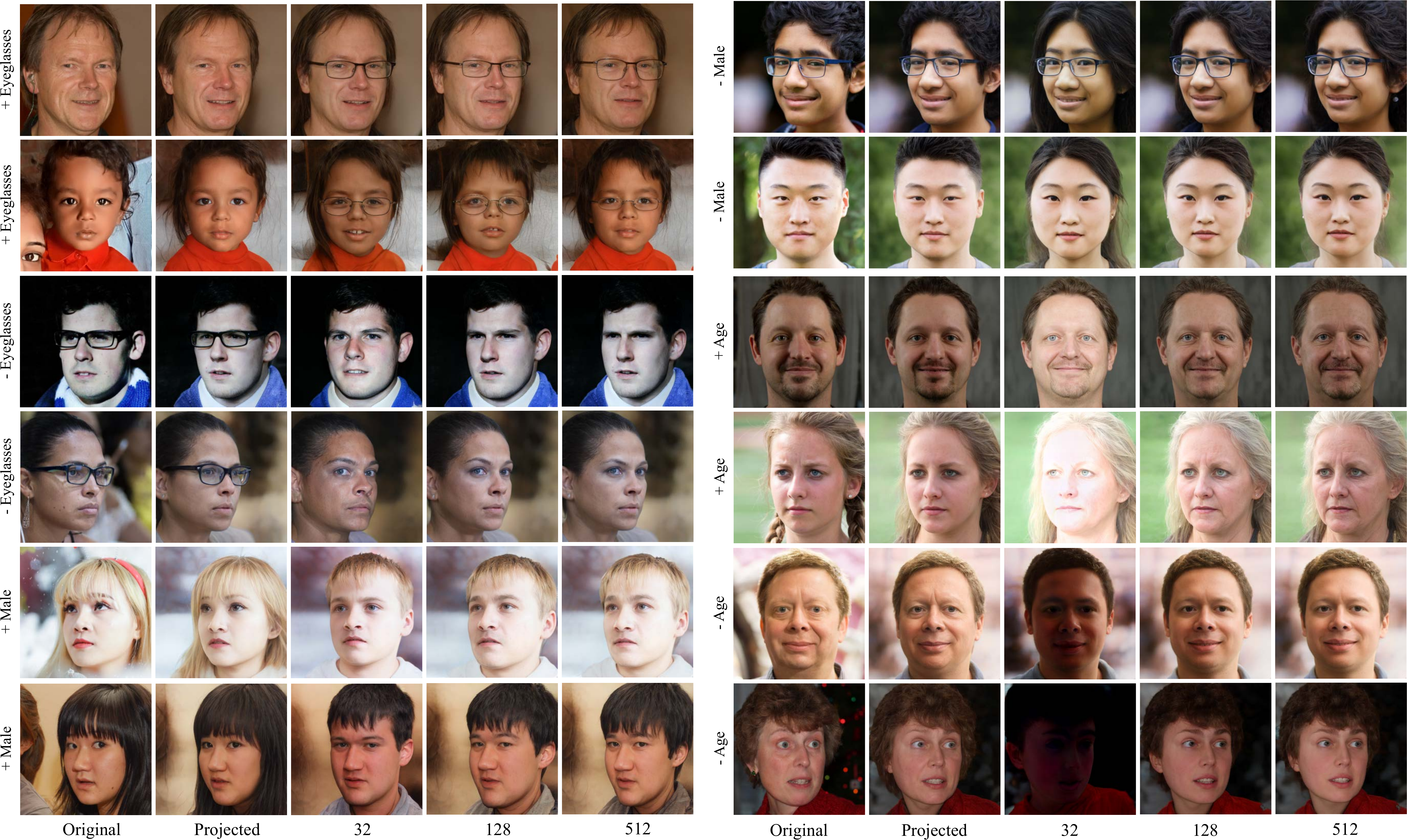}
    \caption{
        Qualitative results for the ablation of the number of labeled samples used for face editing. With only 128 labeled samples, AdaTrans can still achieve satisfactory face editing while 32 samples failed.
    }
    \label{fig:supp_different_labeled_data}
\end{figure*}

Fig.~\ref{fig:ablation_labeled_data} shows the quantitative results for this ablation. Interestingly, we found that training on 128 labeled samples achieves satisfactory performance versus 512 samples. \methodname failed with only 32 samples. Therefore, 128 samples are sufficient to train a good \methodname. Please refer to Fig.~\ref{fig:supp_different_labeled_data} for additional qualitative results.

\section{Additional qualitative results.}

\begin{figure}[h]
    \centering
    \includegraphics[width=1.0\linewidth]{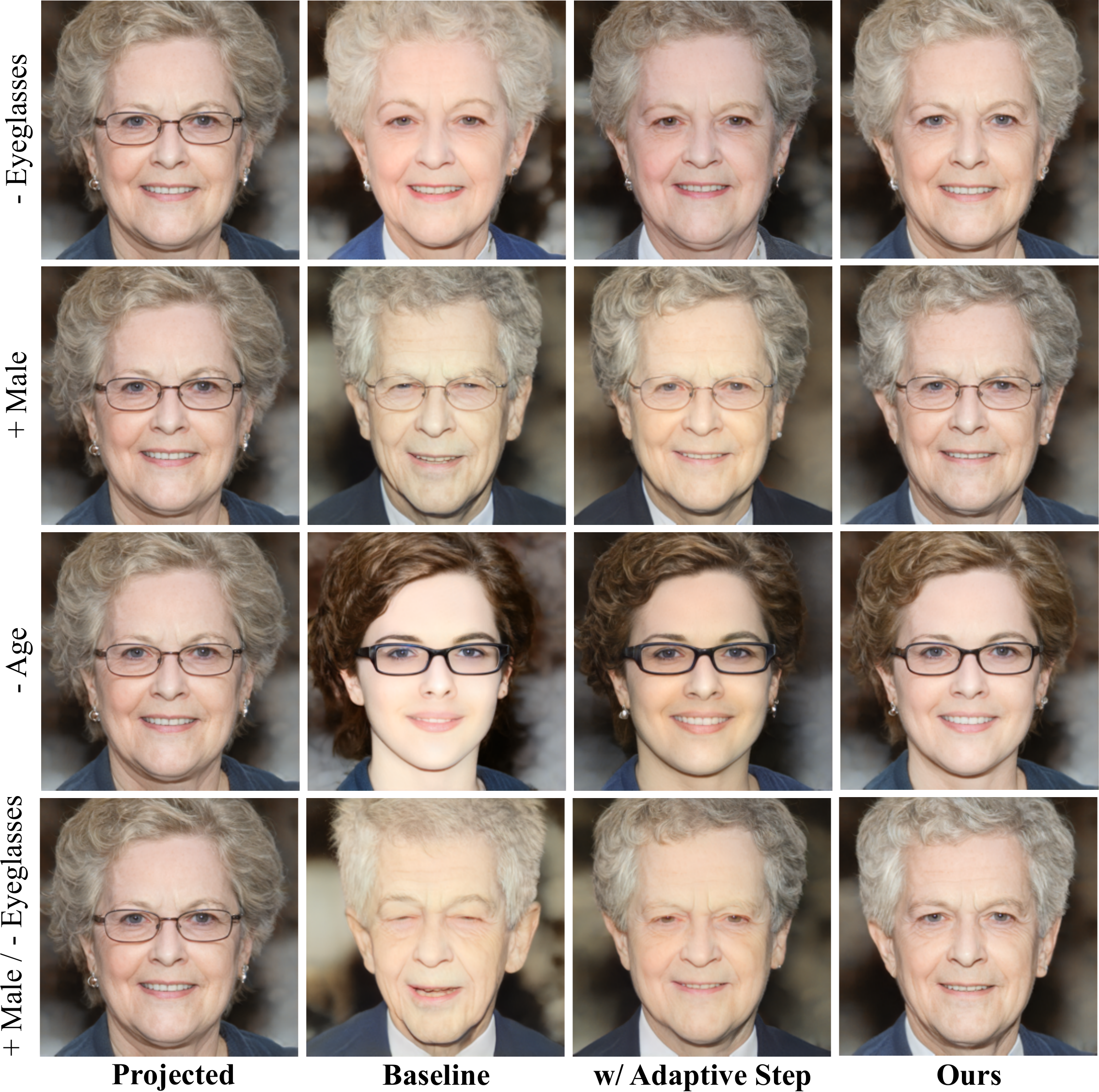}
    \caption{
        Qualitative results for the ablation of the proposed components in \methodname.
        The baseline has inevitably modified some unrelated facial attributes~(\eg, eyes close when editing male and eyeglasses), although the faces are successfully manipulated into target attributes.
        Thanks to the proposed latent regularization, the resultant faces are much more natural than previous two variants.
    }
    \label{fig:ablation_qualitative}
\end{figure}

\begin{figure}[h]
    \centering
    \includegraphics[width=1.0\linewidth]{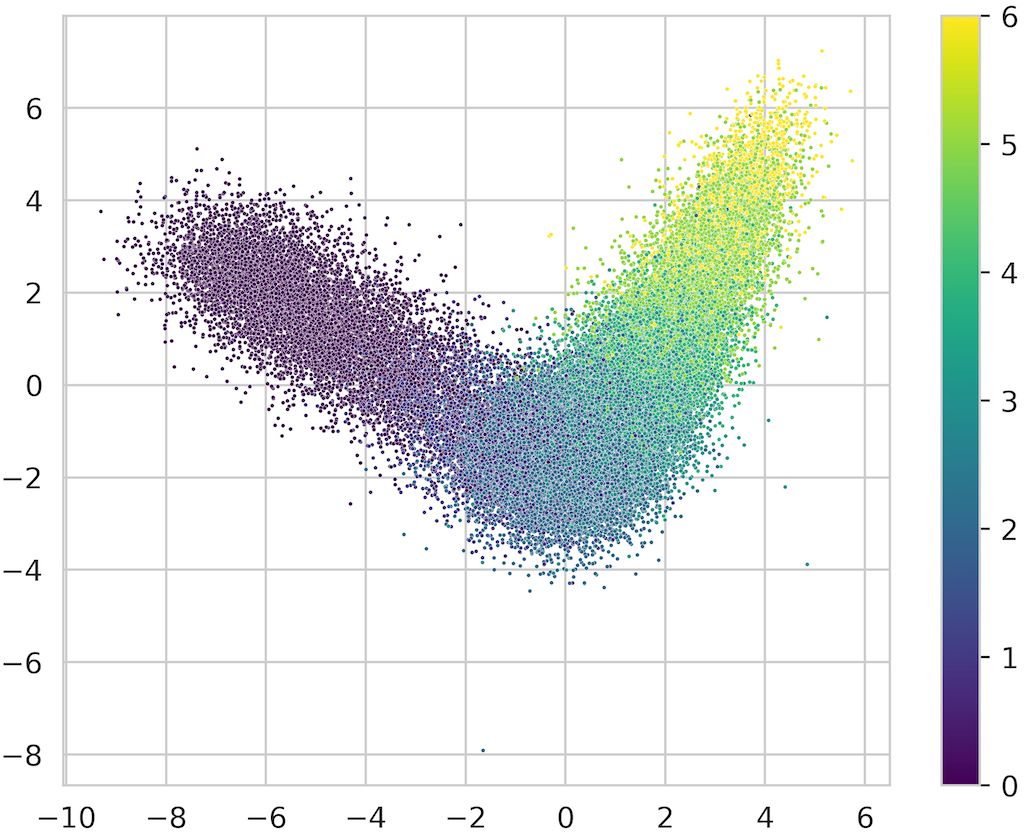}
    \caption{
        Visualization of 100k synthesized latent codes produced by StyleGAN2 mapping network, which is reduced to 2 dimensions by linear discriminant analysis. The points are colored by the age labels obtained using a pre-trained age classifier with 7 discrete age groups~\cite{or2020lifespan}. It validates that the transformation with finer ages cannot be accurately achieved by simple linear interpolation.
    }
    \label{fig:scatter_point_age}
\end{figure}

\begin{figure*}[t]
    \centering
    \includegraphics[width=1.0\linewidth]{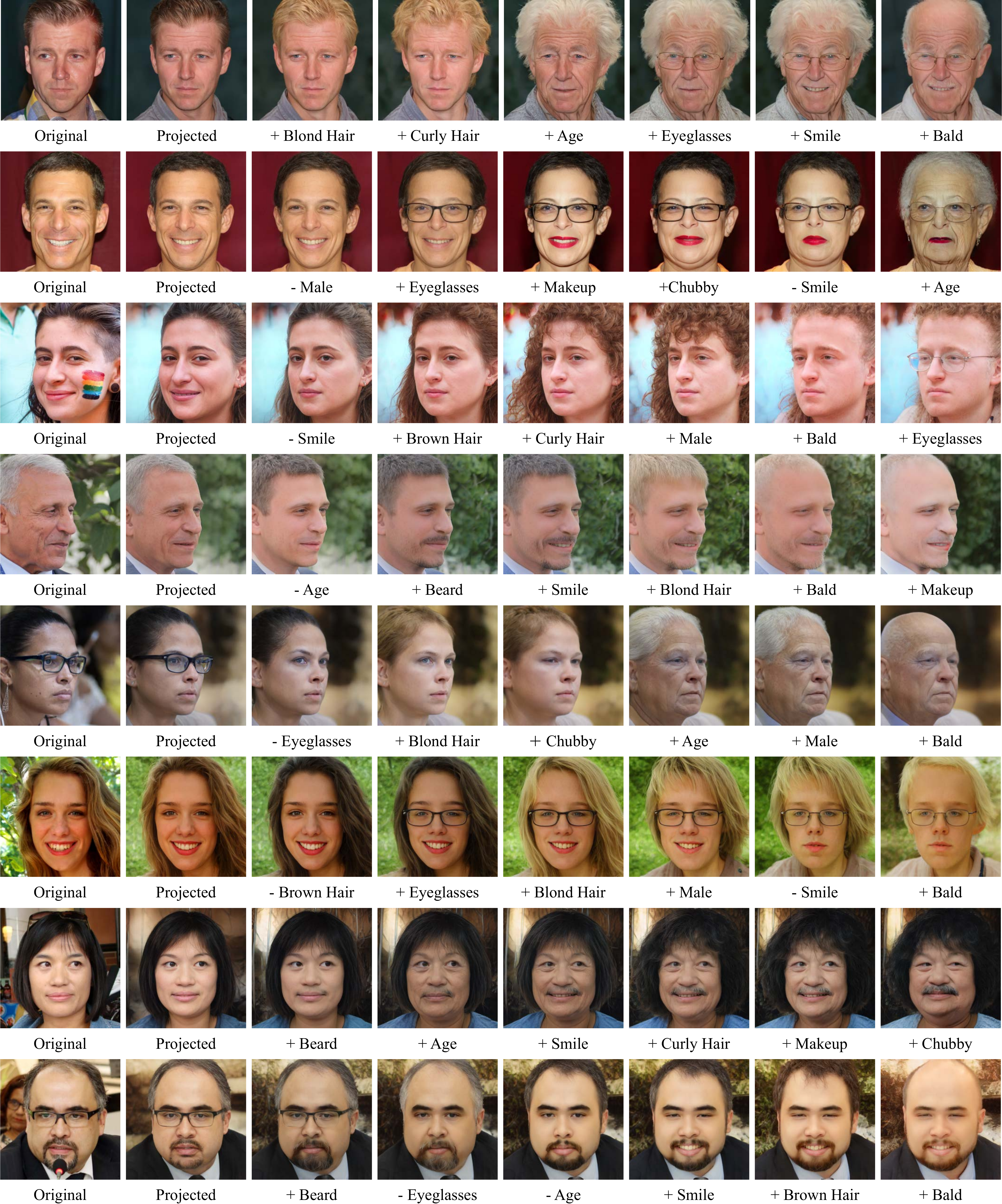}
    \caption{
        Additional results for sequential editing.
    }
    \label{fig:supp-sequence}
\end{figure*}

\begin{figure*}[t]
    \centering
    \includegraphics[width=1.0\linewidth]{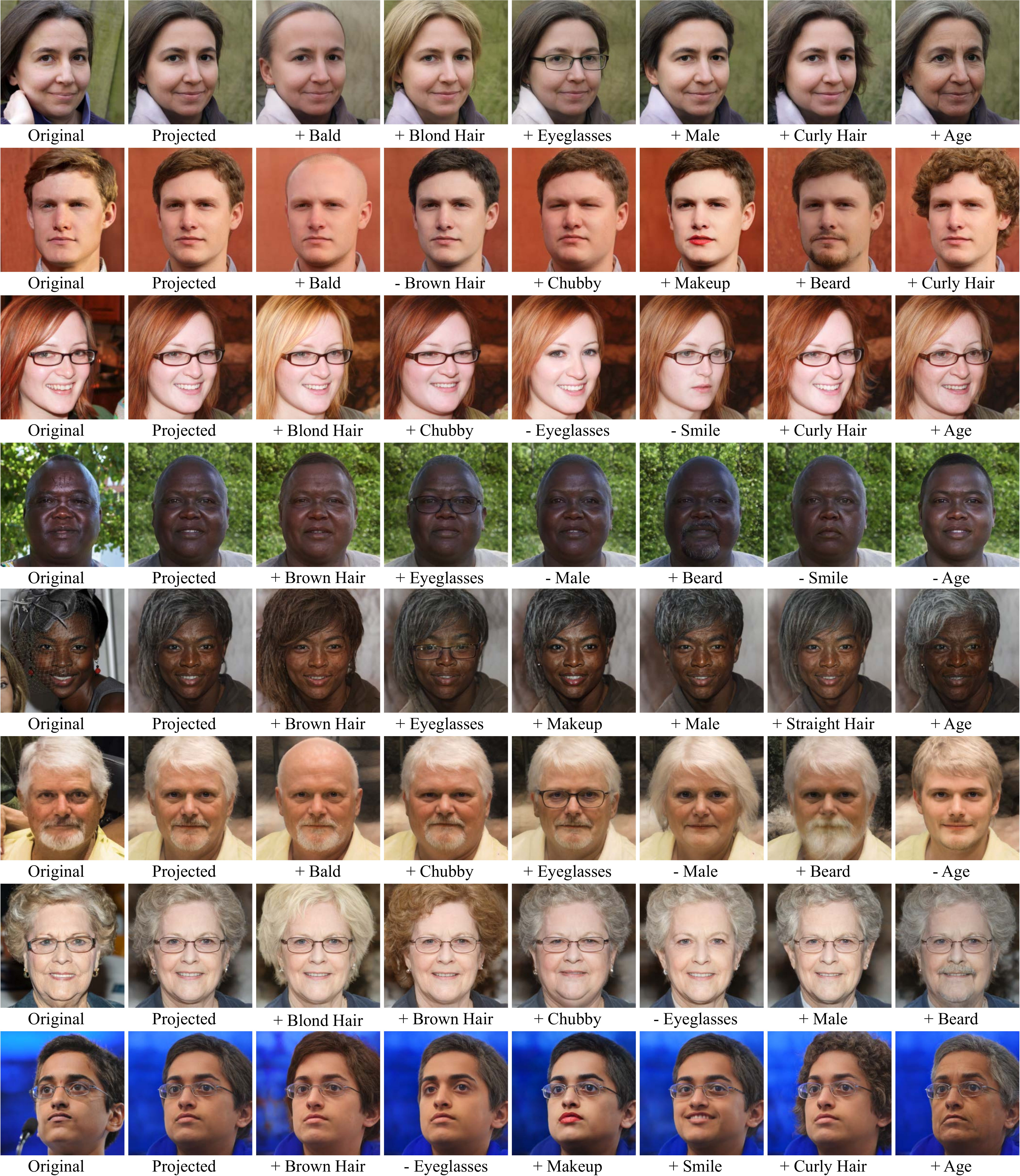}
    \caption{
        Additional results for editing a single attribute.
    }
    \label{fig:supp_single_attribute_editing}
\end{figure*}

\clearpage

\begin{figure*}[t]
	\centering
	\subfloat[Editing single attributes.]{
        \label{fig:qualitative_comparisons_single_attributes_supp}
		\includegraphics[width=.45\textwidth]{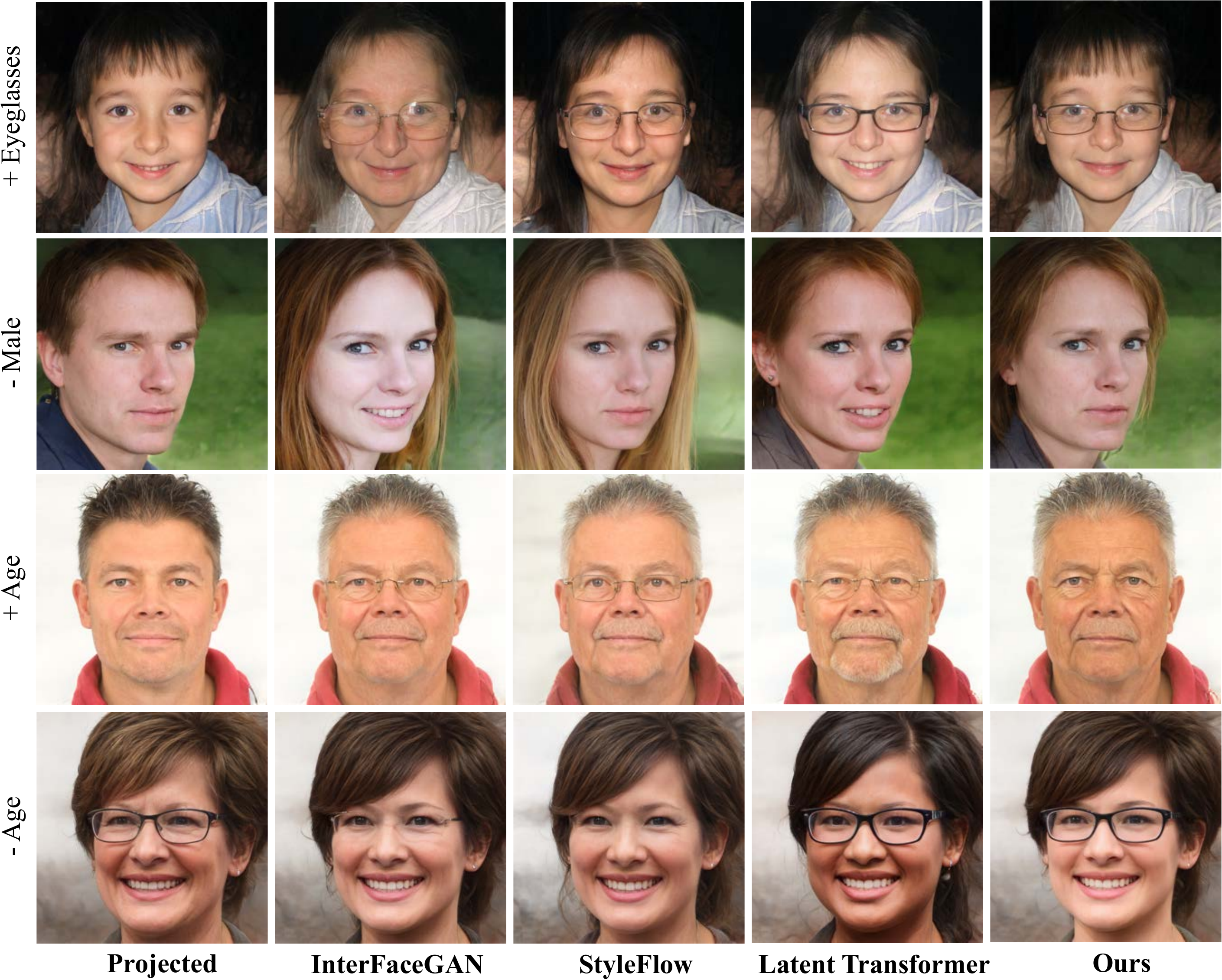}
		}
		\vspace{-10pt}
	\subfloat[Editing multiple attributes.]{
        \label{fig:qualitative_comparisons_multi_attributes_supp}
		\includegraphics[width=.45\textwidth]{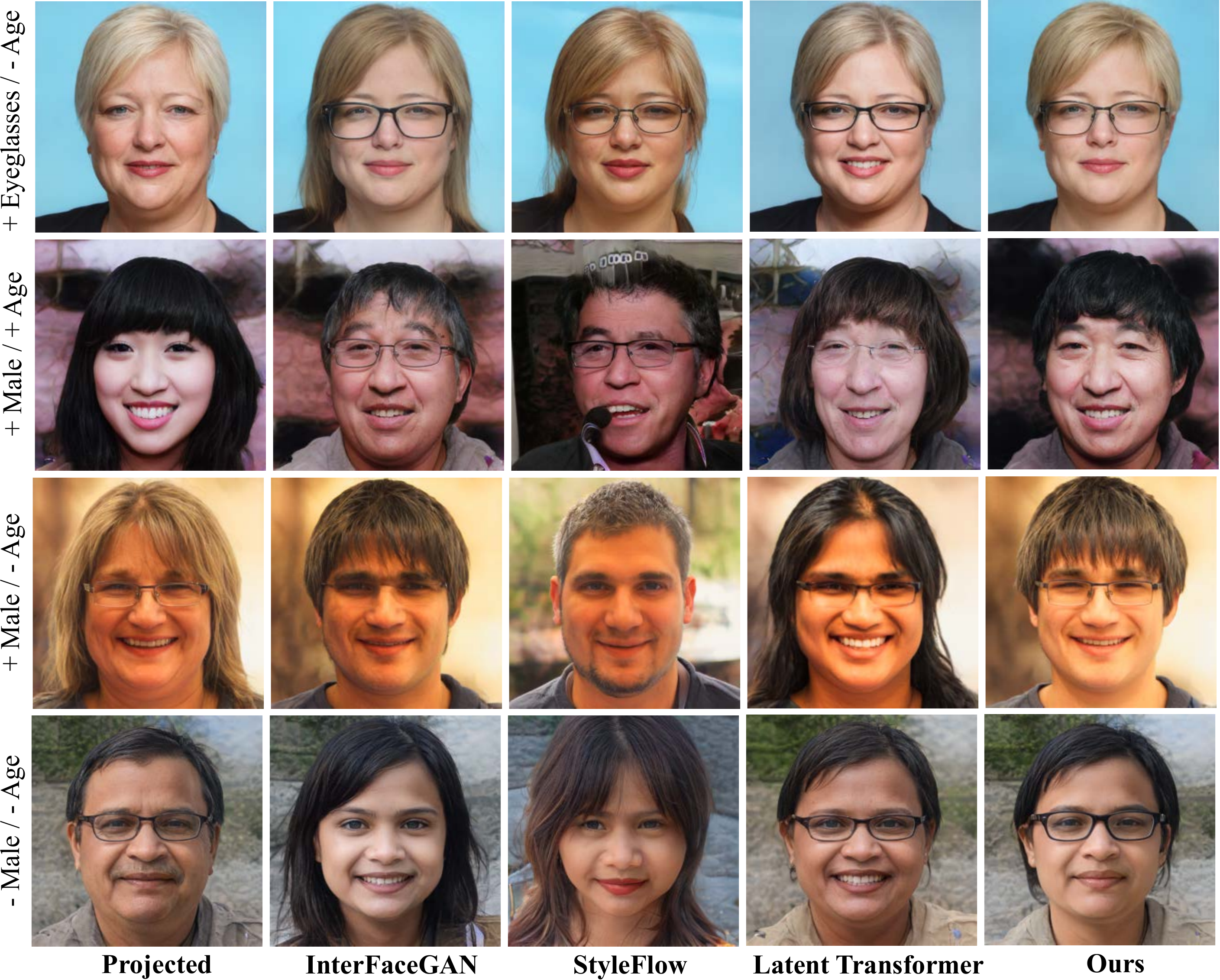}
		}\\
	\subfloat[Editing with large age gap.]{
        \label{fig:qualitative_comparisons_age_gap_supp}
		\includegraphics[width=.45\textwidth]{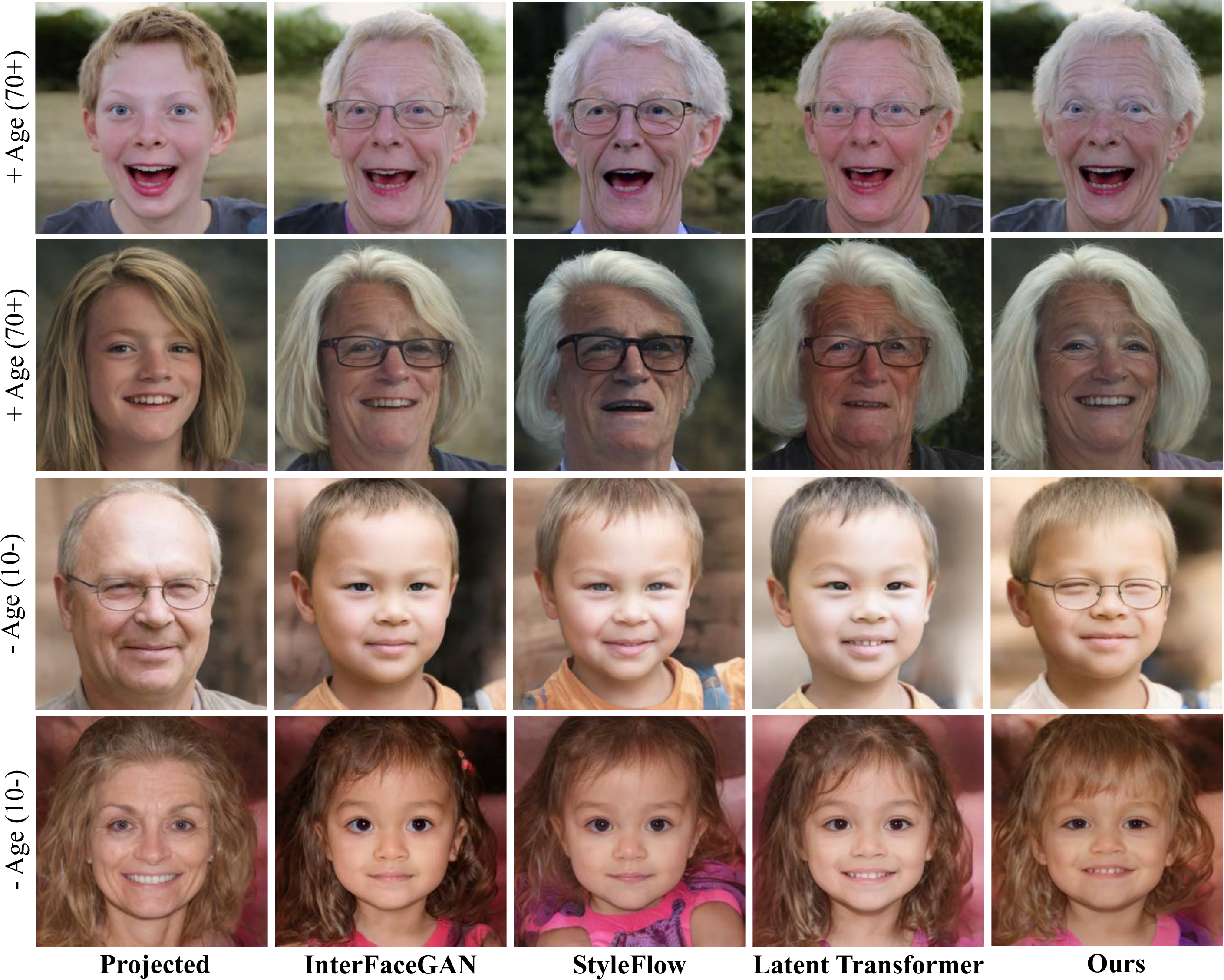}
		}
	\subfloat[Editing with limited labeled data.]{
        \label{fig:qualitative_comparisons_limited_data_single_attributes_supp}
		\includegraphics[width=.45\textwidth]{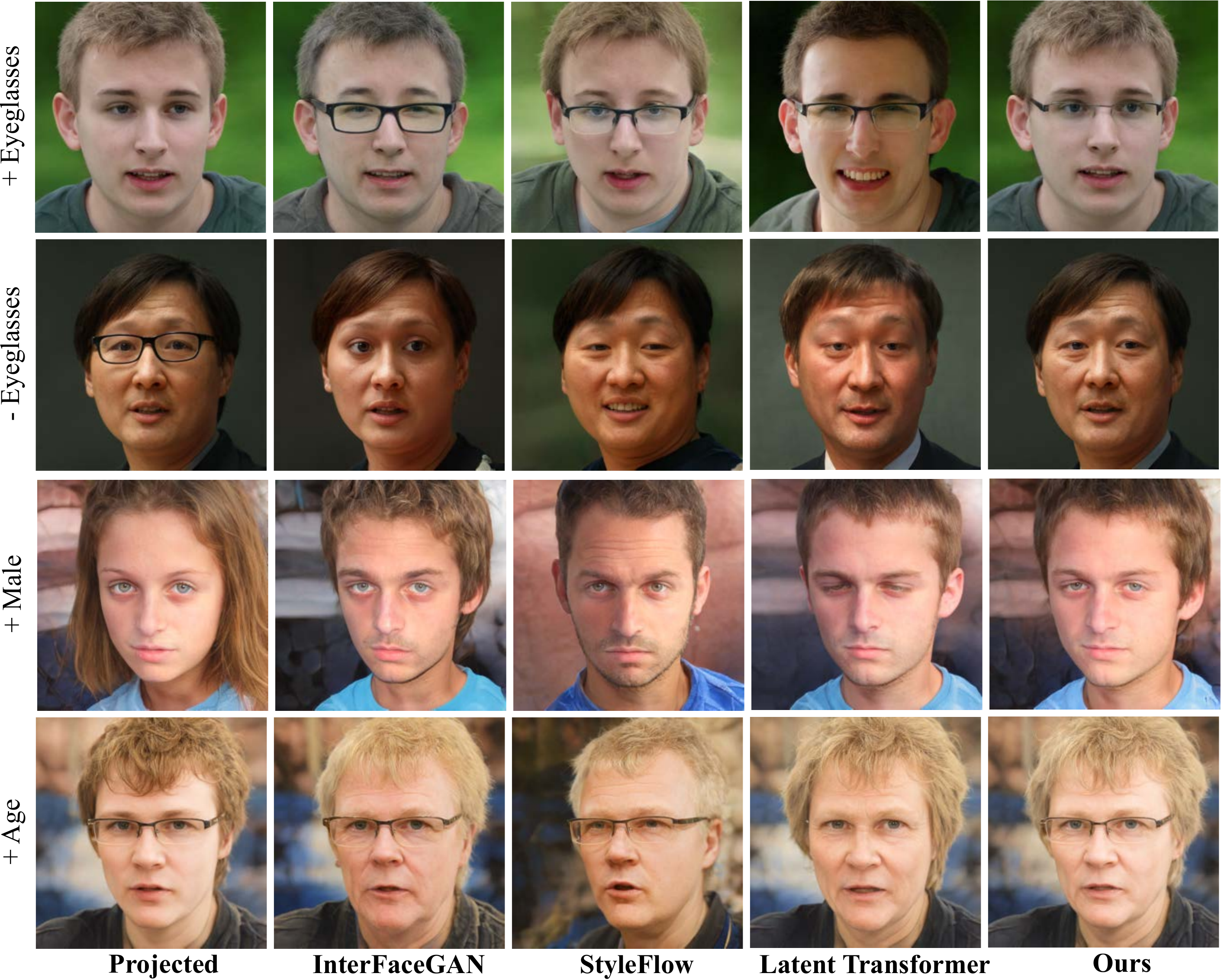}
		}
	\caption{
        Additional results for the qualitative comparisons with recent methods for face editing under different experimental settings.
    }
    \label{fig:supp_qualitative_comparisons}
\end{figure*}

\begin{figure*}[t]
    \centering
    \includegraphics[width=1.0\linewidth]{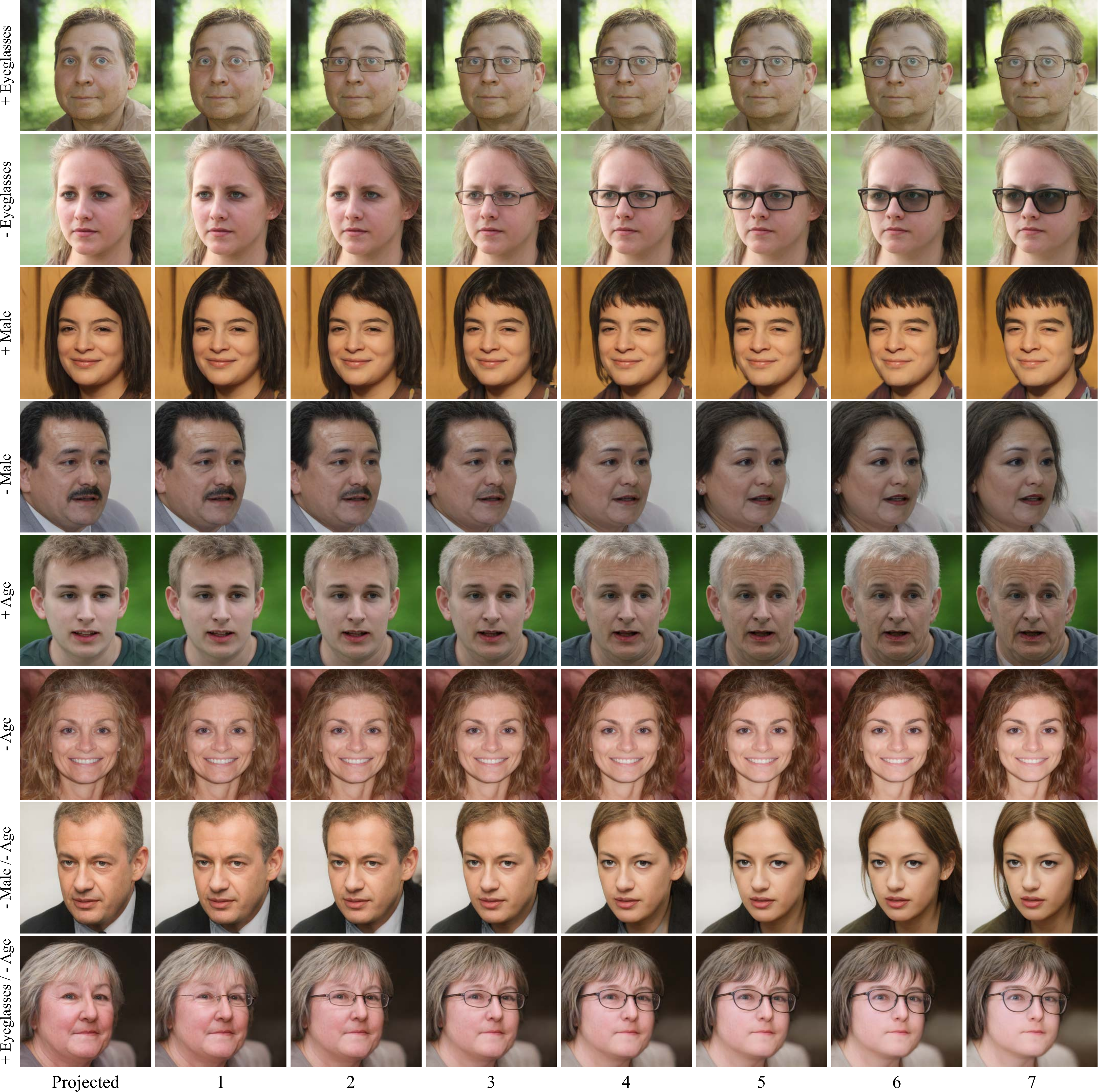}
    \caption{
        Visualization of the intermediate states for AdaTrans, with the step below. Although the maximum step is predefined as 5, AdaTrans can produce photorealistic results at different steps, and would not change unrelated attributes and identities when increasing the maximum step.
    }
    \label{fig:supp_different_step}
\end{figure*}

\end{document}